\documentclass[journal]{IEEEtran}

\usepackage[utf8]{inputenc}
\usepackage{color}
\usepackage{xcolor}
\usepackage{array}
\usepackage{verbatim}
\usepackage{float}
\usepackage{amsmath}
\usepackage{amsthm}
\usepackage{amssymb}
\usepackage{graphicx}
\usepackage{longtable}
\usepackage{multirow}
\usepackage{booktabs}
\usepackage[unicode=true,
bookmarks=false,
breaklinks=false,pdfborder={0 0 1},colorlinks=false]
{hyperref}
\hypersetup{
	colorlinks,bookmarksopen,bookmarksnumbered,citecolor=blue,urlcolor=blue}
\usepackage{cite}

\usepackage{lipsum}
\usepackage{mathtools}
\usepackage{cuted}
\providecommand{\tabularnewline}{\\}
\usepackage{algorithmic}
\usepackage{longtable}

\usepackage{balance}

\usepackage{import}
\usepackage{tikz}
\usepackage{color}
\usepackage{xcolor}
\usepackage{multicol}
\usepackage{array}
\usepackage{verbatim}
\usepackage{amsthm}
\usepackage{longtable}
\usepackage{multirow}
\usepackage{overpic}
\usetikzlibrary{quotes,arrows.meta}
\usetikzlibrary{positioning}

\usepackage{Ball}
\usepackage{Box}
\usepackage{RightBandedBox}

\usetikzlibrary{positioning}
\usetikzlibrary{3d} 
\usepackage[export]{adjustbox}
\usepackage{caption}
\usepackage{dblfloatfix}

\floatstyle{ruled}
\newfloat{algorithm}{tbp}{loa}
\providecommand{\algorithmname}{Algorithm}
\floatname{algorithm}{\protect\algorithmname}

\makeatletter
\let\oldforeign@language\foreign@language
\DeclareRobustCommand{\foreign@language}[1]{%
	\lowercase{\oldforeign@language{#1}}}

\let\oldforeign@language\foreign@language
\DeclareRobustCommand{\foreign@language}[1]{%
	\lowercase{\oldforeign@language{#1}}}

\ifCLASSINFOpdf
\else
\fi

\@ifundefined{showcaptionsetup}{}{%
	\PassOptionsToPackage{caption=false}{subfig}}
\usepackage{subfig}

\usepackage{balance}

\ifCLASSINFOpdf
\else
\fi

\ifCLASSINFOpdf
\else
\fi

\def\ps@IEEEtitlepagestyle{%
	\def\@oddhead{\parbox[t][\height][t]{\textwidth}{\centering \scriptsize
			Personal use of this material is permitted. Permission from the author(s) and/or copyright holder(s), must be obtained for all other uses. Please contact us and provide details if you believe this document breaches copyrights.\\
			\noindent\makebox[\linewidth]{}
		}\hfil\hbox{}}%
	\def\@evenhead{\scriptsize\thepage \hfil \leftmark\mbox{}}%
	\def\@oddfoot{\parbox[t][\height][l]{\textwidth}{
			\vspace{-20pt}{\rule{\textwidth}{0.4pt}}\\ \footnotesize{\bf{\footnotesize\textcolor{red}{K. Ghanizadegan and H. A. Hashim, "DeepUKF-VIN: Adaptively-tuned Deep Unscented Kalman Filter for 3D Visual-Inertial Navigation based on IMU-Vision-Net," Expert Systems With Applications, vol. 271, pp. 126656, 2025.}}} doi: \href{https://doi.org/10.1016/j.eswa.2025.126656}{10.1016/j.eswa.2025.126656}\\
			\noindent\makebox[\linewidth]
		}\hfil\hbox{}}%
	\def\@evenfoot{\MYfooter}}

\makeatother
\begin{document}
	\bstctlcite{IEEEexample:BSTcontrol}

\title{DeepUKF-VIN: Adaptively-tuned Deep Unscented Kalman Filter for 3D Visual-Inertial Navigation based on IMU-Vision-Net}

\author{Khashayar Ghanizadegan and Hashim A. Hashim
\thanks{This work was supported in part by National Sciences and Engineering Research Council of Canada (NSERC), under the grants RGPIN-2022-04937 and DGECR-2022-00103.} 
	\thanks{K. Ghanizadegan and H. A. Hashim are with the Department of Mechanical
		and Aerospace Engineering, Carleton University, Ottawa, Ontario, K1S-5B6,
		Canada (e-mail: hhashim@carleton.ca).}
}



\maketitle
\begin{abstract}
This paper addresses the challenge of estimating the orientation,
position, and velocity of a vehicle operating in three-dimensional
(3D) space with six degrees of freedom (6-DoF). A Deep Learning-based
Adaptation Mechanism (DLAM) is proposed to adaptively tune the noise
covariance matrices of Kalman-type filters for the Visual-Inertial
Navigation (VIN) problem, leveraging IMU-Vision-Net. Subsequently,
an adaptively tuned Deep Learning Unscented Kalman Filter for 3D VIN
(DeepUKF-VIN) is introduced to utilize the proposed DLAM, thereby
robustly estimating key navigation components, including orientation,
position, and linear velocity. The proposed DeepUKF-VIN integrates
data from onboard sensors, specifically an inertial measurement unit
(IMU) and visual feature points extracted from a camera, and is applicable
for GPS-denied navigation. Its quaternion-based design effectively
captures navigation nonlinearities and avoids the singularities commonly
encountered with Euler-angle-based filters. Implemented in discrete
space, the DeepUKF-VIN facilitates practical filter deployment. The
filter's performance is evaluated using real-world data collected
from an IMU and a stereo camera at low sampling rates. The results
demonstrate filter stability and rapid attenuation of estimation errors,
highlighting its high estimation accuracy. Furthermore, comparative
testing against the standard Unscented Kalman Filter (UKF) in two
scenarios consistently shows superior performance across all navigation
components, thereby validating the efficacy and robustness of the
proposed DeepUKF-VIN.
\end{abstract}


\rule{0.46\textwidth}{1pt}\\
For video of navigation experiment visit: \href{https://youtu.be/japfpySxilA}{link}
\\
\vspace{-1pt}
\rule{0.49\textwidth}{1pt}

\section{Introduction}\label{sec1}

\subsection{Motivation}
\IEEEPARstart{N}{avigation} is a fundamental component in the successful operation
of a wide array of applications, spanning fields such as robotics,
aerospace, and mobile technology. At its core, navigation involves
estimating an object's position, orientation, and velocity, a task
that becomes particularly critical and challenging in environments
where Global Navigation Satellite Systems (GNSS), like GPS, BeiDou,
and GLONASS, are unavailable (e.g., indoor environments) or unreliable
(e.g., urban settings with obstructed satellite signals due to tall
buildings) \cite{hash2025_RIENG_Avionics,hashim2021geometricNAV}.
Similar challenges are encountered in underwater navigation, where
robots must operate in deep, GNSS-denied environments \cite{zhang2023underwater}.
Unmanned ground vehicles (UGVs) and unmanned aerial vehicles (UAVs)
have shown immense potential in various sectors. For example, UGVs
and UAVs are increasingly used in care facilities to assist with monitoring
and delivery tasks \cite{yang2020homecare}, in logistical services
for autonomous package delivery \cite{saunders2024autonomous}, and
in surveillance of hard-to-access locations \cite{hash2025_RIENG_Avionics}.
These include monitoring forests for early fire detection \cite{hu2022fault},
tracking icebergs in the Arctic \cite{kaiser2022potential}, and conducting
surveys in other remote areas. The effectiveness of these applications
hinges on the precision and reliability of their navigation systems.
In the realm of mobile technology, accurate navigation is essential
for enhancing user experiences, particularly in smartphone applications
that rely on real-time positional data, such as augmented reality
(AR) platforms and wayfinding tools \cite{chen2021augmented}. Similarly,
in aerospace applications, obtaining precise positional and orientation
data is vital for the accurate analysis and interpretation of observational
information \cite{korkin2024multiparticle,aerospace7010003}.

\subsection{Related Work}

One of the primary approaches to addressing the challenge of navigation
in GPS-denied environments involves utilizing ego-acceleration measurements
from onboard accelerometers to estimate a vehicle's pose relative
to its previous position. This technique, known as Dead Reckoning
(DR), integrates acceleration data to derive positional information
\cite{hash2025_RIENG_Avionics}. DR offers a straightforward, cost-effective
solution, particularly with low-cost sensors, making it accessible
for many applications \cite{Hou9162069Pedestrian}. To enhance the
accuracy of Dead Reckoning, a gyroscope is often incorporated to measure
the vehicle's angular velocity. This integration provides additional
orientation data, improving the overall pose estimation. However,
a significant drawback of this method is its susceptibility to cumulative
errors or drift over time. Without supplementary sensors or correction
mechanisms, these errors can accumulate rapidly, leading to inaccurate
navigation results, especially during prolonged use. In controlled
environments like harbors, warehouses, or other predefined spaces,
ultra-wideband (UWB) technology can significantly enhance navigation
accuracy. UWB systems measure distances between the vehicle and fixed
reference points, known as anchors, providing highly accurate and
robust localization data \cite{Hashim2023NonlinearFusion}. This approach
is widely adopted in applications where precision is paramount, such
as robotic operations within structured environments and object-tracking
systems like Apple's AirTag \cite{Roth2022AirTag}. When used alongside
an Inertial Measurement Unit (IMU), accelerometer, and gyroscope within
a DR framework, UWB can serve as an additional sensor to correct positional
errors and mitigate drift \cite{Hashim2023NonlinearFusion}. However,
this solution has limitations since UWB requires the installation
of anchors in the environment, which confines its applicability to
pre-configured spaces. As a result, it may not be suitable for dynamic
or unstructured environments, reducing the system's flexibility and
immediate usability out of the box \cite{hashim2023exponentially}.
Additionally, UWB is susceptible to high levels of noise, which can
degrade estimation accuracy \cite{Hashim2023NonlinearFusion}.

With the development of advanced point cloud registration algorithms
such as Iterative Closest Point (ICP) \cite{121791} and Coherent
Point Drift (CPD) \cite{Myronenko2009PointSR}, sensors capable of
capturing two-dimensional (2D) points from three-dimensional (3D)
space have emerged as promising candidates to complement IMUs without
requiring prior environmental knowledge. Sound Navigation and Ranging
(SONAR) is one such sensor, widely adopted in marine applications
due to its effectiveness in underwater environments, where mechanical
waves propagate efficiently \cite{franchi2021underwater}. Similarly,
Light Detection and Ranging (LiDAR) employs electromagnetic waves
instead of mechanical waves and has demonstrated utility in aerospace
applications, where sound propagation is limited, but light transmission
is effective \cite{christian2013survey}. However, both SONAR and
LiDAR exhibit significant limitations in complex indoor and outdoor
environments, as they rely solely on structural properties and cannot
capture texture or color information. In contrast, recent advancements
in low-cost, high-resolution cameras designed for navigation applications,
combined with robust fusion between IMU and feature detection \cite{hashim2021geometricNAV,hashim2021gps}.
Popular tracking feature detection-based algorithms include Scale-Invariant
Feature Transform (SIFT) \cite{lowe2004distinctive}, Good Features
to Track (GFTT) \cite{shi1994good}, and the Kanade-Lucas-Tomasi (KLT)
algorithm have facilitated the widespread adoption of cameras as correction
sensors alongside IMUs \cite{7511659,mourikis2007multi,sun2018robust,hash2025_RIENG_Avionics}.

\subsection{Persistent Challenges and Potentials}

To integrate the aforementioned sensor data, Kalman-type filters are
widely employed in navigation due to their stochastic framework and
ability to handle noisy measurements \cite{odry2018kalman,Khashayar2025_TIM_Hashim,hashim2019Ito,hashim2020systematic}.
The Kalman Filter (KF) provides a maximum likelihood estimate of the
system's state vector based on available measurement data; however,
it operates optimally only within linear systems. To overcome this
limitation, the Extended Kalman Filter (EKF) was developed. The EKF
linearizes the system around the current estimated state vector and
applies the KF framework to this linearized model. Its intuitive structure,
ease of implementation, and computational efficiency have established
the EKF as a standard choice for navigation applications \cite{Myronenko2009PointSR,Erdem2015FusingTracking,mourikis2007multi}.
However, the EKF’s performance degrades with increasing system nonlinearity.
To address the EKF's limitations, the Unscented Kalman Filter (UKF)
was introduced. The UKF effectively captures the propagation of mean
and covariance through a nonlinear transformation up to the second
order, offering improved accuracy while maintaining comparable computational
complexity to the EKF \cite{ref:ukf,Khashayar2025_TIM_Hashim}. Nevertheless,
Kalman-type filters rely on accurate modeling of system and measurement
noise. While it is standard to assume these noise components are zero-mean,
their covariance matrices serve as critical tuning parameters, and
the performance of these filters is sensitive to inaccuracies in their
specification \cite{wernitz2022noise}. In practice, determining the
values of covariance matrices is challenging and typically involves
an iterative process of trial and error, which can be both time-consuming
and effort-intensive.

Deep learning techniques have shown significant promise in adaptively
tuning the covariance matrices of Kalman-based filters, addressing
a critical challenge in achieving accurate state estimation \cite{brossard2020ai,Zhou2022IMUAlgorithm,or2023learning,yan2023multi}.
These methods offer an efficient alternative to traditional manual
tuning, leveraging data-driven models to dynamically estimate noise
parameters based on observed system behavior. For instance, Brossard
et al. \cite{brossard2020ai} utilized Convolutional Neural Networks
(CNNs) to predict measurement noise parameters for the DR of ground
vehicles using an Invariant EKF (IEKF). This approach improved noise
estimation by learning from raw sensor data, enhancing overall navigation
accuracy. Similarly, Or et al. \cite{or2023learning} applied deep
learning to model trajectory uncertainty by extracting features such
as vehicle speed and path curvature demonstrating the potential to
enhance state predictions by accurately capturing the system's dynamic
characteristics. Furthermore, Yan et al. \cite{yan2023multi} proposed
a multi-level framework where the state vector estimates and covariance
predictions from traditional filters serve as inputs to deep learning
architectures. Therefore, deep learning can be employed to iteratively
refine the covariance estimates, improving robustness in complex scenarios
allowing Kalman-based filters to dynamically adjust varying noise
conditions, reducing dependency on intensive manual tuning and significantly
improving performance in real-world applications.

\subsection{Contributions}

Motivated by the above discussion, the key contributions of this work
are as follows: (1) The proposed approach employs singularity-free
quaternion dynamics to represent ego orientation, ensuring robust
handling of orientation estimation and avoiding singularities typically
encountered with Euler-angle-based representations. (2) A quaternion-based,
adaptively-tuned Deep Learning Unscented Kalman Filter for 3D Visual-Inertial
Navigation (DeepUKF-VIN) based on Deep Learning-based Adaptation Mechanism
(DLAM) is formulated in discrete form. This approach accurately models
the true navigation kinematics, simplifies the implementation process,
and dynamically estimates the covariance matrices, thereby enhancing
the overall performance of Kalman-type filters. (3) A novel deep learning-based
adaptation mechanism is introduced to dynamically estimate the covariance
matrices associated with the measurement noise vectors in the UKF.
This adaptive approach enhances the filter's estimation performance
by reducing dependency on manual tuning. (4) The proposed DeepUKF-VIN
demonstrates superior performance compared to the standard UKF across
various scenarios. DeepUKF-VIN effectiveness is validated using real-world
data collected from low-cost sensors operating at low sampling rates.
To the best of the authors' knowledge, no deep learning-enhanced Kalman-type
filter based on inertial measurement and vision units has been proposed
for VIN.

\subsection{Structure}

The structure of the paper is organized as follows: Section \ref{sec:Preliminaries-and-Math}
introduces the preliminary concepts and mathematical foundations.
Section \ref{sec:Problem-Formulation} defines the nonlinear navigation
kinematics problem. Section \ref{sec:QNUKF} presents the quaternion-based
UKF framework tailored for navigation kinematics. Section \ref{sec:DLAM}
provides a detailed description of the deep learning architecture
for adaptive tuning. Section \ref{sec:Train} outlines the training
process and implementation methodology of the proposed DeepUKF-VIN.
Section \ref{sec:validation} evaluates the performance of the DeepUKF-VIN
algorithm using a real-world dataset. Finally, Section \ref{sec:Conclusion}
offers concluding remarks.

\section{Preliminaries\label{sec:Preliminaries-and-Math}}

\begin{table}[t]
	\centering{}\caption{\label{tab:Table-of-Notations2}Nomenclature}
	\begin{tabular}{>{\raggedright}p{2cm}l>{\raggedright}p{5.6cm}}
		\toprule 
		\addlinespace
		$\left\{ \mathcal{B}\right\} $ / $\left\{ \mathcal{W}\right\} $  & :  & Fixed body-frame / fixed world-frame\tabularnewline
		\addlinespace
		$\mathbb{SO}\left(3\right)$  & :  & Special Orthogonal Group of order 3\tabularnewline
		\addlinespace
		$\mathbb{S}^{3}$  & :  & Three-unit-sphere\tabularnewline
		\addlinespace
		$q_{k},\hat{q}_{k}$  & :  & True and estimated quaternion at step $k$\tabularnewline
		\addlinespace
		$p_{k},\hat{p}_{k}$  & :  & True and estimated position at step $k$\tabularnewline
		\addlinespace
		$v_{k},\hat{v}_{k}$  & :  & True and estimated linear velocity at step $k$\tabularnewline
		\addlinespace
		$r_{e,k}$, $p_{e,k}$, $v_{e,k}$  & :  & Attitude, position, and velocity estimation error\tabularnewline
		\addlinespace
		$a_{k},a_{m,k}$  & :  & True and measured acceleration at step $k$\tabularnewline
		\addlinespace
		$\omega_{k},\omega_{m,k}$  & :  & True and measured angular velocity at step $k$\tabularnewline
		\addlinespace
		$\eta_{\omega,k},\eta_{a,k}$  & :  & Angular velocity and acceleration measurements noise\tabularnewline
		\addlinespace
		$b_{\omega,k},b_{a,k}$  & :  & Angular velocity and acceleration measurements bias\tabularnewline
		\addlinespace
		$C_{\times}$  & :  & Covariance matrix of $n_{\times}$.\tabularnewline
		\addlinespace
		$l_{b,k},l_{b,w}$  & :  & landmark coordinates in $\left\{ \mathcal{B}\right\} $ and $\left\{ \mathcal{W}\right\} $.\tabularnewline
		\addlinespace
		$x_{k}$, $x_{k}^{a}$, $u_{k}$  & :  & The state, augmented state, and input vectors at the $k$th time step\tabularnewline
		\addlinespace
		$\hat{z}_{k},z_{k}$  & :  & Predicted and true measurement\tabularnewline
		\addlinespace
		$\{\chi_{i|j}\}_{\nu}$, $\{\chi_{i|j}^{a}\}_{\nu}$, $\{\zeta_{i|j}\}_{\nu}$  & :  & Sigma points of state, augmented state, and measurements\tabularnewline
		\bottomrule
	\end{tabular}
\end{table}

\paragraph*{Notation}In this paper, the set of $d_{1}$-by-$d_{2}$
matrices of real numbers is denoted by $\mathbb{R}^{d_{1}\times d_{2}}$.
A vector $v\in\mathbb{R}^{d}$ is said to lie on the $d$-dimensional
sphere $\mathbb{S}^{d-1}\subset\mathbb{R}^{d}$ when its norm, denoted
as $\|m\|=\sqrt{m^{\top}m}\in\mathbb{R}$, is equal to one. The identity
matrix of dimension $d$ is denoted by $\mathbf{I}_{d}\in\mathbb{R}^{d\times d}$.
The world frame $\left\{ \mathcal{W}\right\} $ and the body frame
$\left\{ \mathcal{B}\right\} $ refer to the coordinate systems attached
to the Earth and the vehicle, respectively. Table \ref{tab:Table-of-Notations2}
lists a summary of notations heavily used in this paper.

\subsection{Preliminary}

The matrix $R\in\mathbb{R}^{3\times3}$ represents the vehicle's orientation,
provided it belongs to the Special Orthogonal Group of order 3, denoted
$\mathbb{SO}(3)$, which is defined by:
\[
\mathbb{SO}(3):=\left\{ \left.R\in\mathbb{R}^{3\times3}\right|det(R)=+1,RR^{\top}=\mathbf{I}_{3}\right\} 
\]
A quaternion vector $q$ is defined in the scalar-first format as
$q=[q_{w},q_{x},q_{y},q_{z}]^{\top}=[q_{w},q_{v}^{\top}]^{\top}\in\mathbb{S}^{3}$
with $q_{v}\in\mathbb{R}^{3}$, $q_{w}\in\mathbb{R}$, and $\mathbb{S}^{3}:=\{\left.q\in\mathbb{R}^{4}\right|||q||=1\}$.
To obtain the quaternion representation corresponding to a rotation
matrix $R=\begin{bmatrix}R_{(1,1)} & R_{(1,2)} & R_{(1,3)}\\
	R_{(2,1)} & R_{(2,2)} & R_{(2,3)}\\
	R_{(3,1)} & R_{(3,2)} & R_{(3,3)}
\end{bmatrix}$, the mapping $q_{R}:\mathbb{SO}(3)\rightarrow\mathbb{S}^{3}$ is
defined as \cite{hashim2019special}:
\begin{align}
	q_{R}(R)=\left[\begin{array}{c}
		q_{w}\\
		q_{x}\\
		q_{y}\\
		q_{z}
	\end{array}\right] & =\left[\begin{array}{c}
		\frac{1}{2}\sqrt{1+R_{(1,1)}+R_{(2,2)}+R_{(3,3)}}\\
		\frac{1}{4q_{w}}(R_{(3,2)}-R_{(2,3)})\\
		\frac{1}{4q_{w}}(R_{(1,3)}-R_{(3,1)})\\
		\frac{1}{4q_{w}}(R_{(2,1)}-R_{(1,2)})
	\end{array}\right]\label{R2Q}
\end{align}
The orientation resulting from two subsequent rotations $q_{1}=[q_{w1},q_{v1}]^{\top}\in\mathbb{S}^{3}$
and $q_{2}=[q_{w2},q_{v2}]^{\top}\in\mathbb{S}^{3}$ is defined through
quaternion multiplication, denoted by the $\otimes$ operator \cite{hashim2019special}:
\begin{align}
	q_{3} & =q_{1}\otimes q_{2}\nonumber \\
	& =\begin{bmatrix}q_{w1}q_{w2}-q_{v1}^{\top}q_{v2}\\
		q_{w1}q_{v2}+q_{w2}q_{v1}+[q_{v1}]_{\times}q_{v2}
	\end{bmatrix}\in\mathbb{S}^{3}\label{eq:qxq}
\end{align}
The orientation identical in terms of unit quaternion is $q_{I}=[1,0,0,0]^{\top}$.
For $q=[q_{w},q_{v}^{\top}]^{\top}\in\mathbb{S}^{3}$, the inverse
of $q$ is given by $q^{-1}=[q_{w},-q_{v}^{\top}]^{\top}\in\mathbb{S}^{3}$.
It is worth noting that $q\otimes q^{-1}=q_{I}$. For $m\in\mathbb{R}^{3}$,
the skew-symmetric matrix $[m]_{\times}$ is defined as:
\[
[m]_{\times}=\left[\begin{array}{ccc}
	0 & -m_{3} & m_{2}\\
	m_{3} & 0 & -m_{1}\\
	-m_{2} & m_{1} & 0
\end{array}\right]\in\mathfrak{so}(3),\hspace{1em}m=\left[\begin{array}{c}
	m_{1}\\
	m_{2}\\
	m_{3}
\end{array}\right]
\]
The mapping from quaternion $q=[q_{w},q_{v}^{\top}]^{\top}\in\mathbb{S}^{3}$
to rotation matrix $R_{q}\in\mathbb{SO}(3)$ is defined by \cite{hashim2019special}:
\begin{equation}
	R_{q}(q)=(q_{w}^{2}-\|q_{v}\|^{2})I_{3}+2q_{v}q_{v}^{\top}+2q_{w}[q_{v}]_{\times}\label{Q2R}
\end{equation}
The inverse of the skew-symmetric matrix functionis given by:
\begin{equation}
	{\rm vex}([m]_{\times})=m\in\mathbb{R}^{3}\label{eq:vex}
\end{equation}
Let $\mathcal{P}_{a}(\cdot):\mathbb{R}^{3\times3}\rightarrow\mathfrak{so}(3)$
be anti-symmetric projection operator where
\begin{equation}
	\mathcal{P}_{a}(M)=\frac{1}{2}(M-M^{\top})\in\mathfrak{so}(3),\hspace{0.5em}\forall M\in\mathbb{R}^{m\times m}\label{eq:pa}
\end{equation}
The orientation of a rigid body can also be represented by a rotation
angle $\theta\in\mathbb{R}$ around a unit vector $u\in\mathbb{S}^{2}\subset\mathbb{R}^{3}$
with $\mathbb{S}^{2}:=\{\left.u\in\mathbb{R}^{3}\right|||u||=1\}$.
Angle-axis parametrization is obtained from the rotation matrix $R\in\mathbb{SO}(3)$,
where \cite{hashim2019special}:
\begin{equation}
	\left\{ \begin{aligned}\theta_{R}(R) & =\arccos\left(\frac{{\rm Tr}(R)-1}{2}\right)\in\mathbb{R}\\
		u_{R}(R) & =\frac{1}{\sin(\theta_{R}(R))}{\rm vex}(\mathcal{P}_{a}(R))\in\mathbb{S}^{2}
	\end{aligned}
	\right.\label{R2AA}
\end{equation}
with ${\rm Tr}(\cdot)$ denoting the trace function. Let the rotation
vector $r$ be described via the angle-axis parametrization as follows:
\begin{equation}
	r=r_{\theta,u}(\theta,u)=\theta u\in\mathbb{R}^{3},\hspace{1em}\forall\theta\in\mathbb{R},\,u\in\mathbb{S}^{2}\label{AA2r}
\end{equation}
The rotation matrix associated with a rotation vector is given by
\cite{hashim2019special}: 
\begin{align}
	R_{r}(r) & =\exp([r]_{\times})\in\mathbb{SO}(3)\label{r2R}
\end{align}
The mapping from rotation vector representation to quaternion representation
is found by utilizing \eqref{R2Q}, \eqref{R2AA}, and \eqref{AA2r}
such that $q_{r}:\mathbb{R}^{3}\rightarrow\mathbb{S}^{3}$: 
\begin{align}
	q_{r}(r) & =q_{R}\left(R_{r}\left(r\right)\right)\in\mathbb{S}^{3}\label{eq:r2q}
\end{align}
The rotation vector corresponding to a rotation represented by a quaternion
is found in light of \eqref{Q2R}, \eqref{R2AA}, and \eqref{AA2r}
by $r_{q}:\mathbb{S}^{3}\rightarrow\mathbb{R}^{3}$ such that
\begin{equation}
	r_{q}(q)=r_{\theta,u}\left(\theta_{R}\left(R_{q}(q)\right),u_{R}\left(R_{q}(q)\right)\right)\label{eq:q2r}
\end{equation}
To facilitate addition $\boxplus$ and subtraction $\boxminus$ between
a rotation vector $r\in\mathbb{R}^{3}$ and a quaternion $q\in\mathbb{S}^{3}$,
using the definitions in \eqref{eq:vex}, \eqref{AA2r}, and \eqref{eq:r2q},
the following operations are defined:
\begin{align}
	q\boxplus r & :=q_{r}(r)\otimes q\in\mathbb{S}^{3}\label{eq:q_p_r}\\
	q\boxminus r & :=q_{r}(r)^{-1}\otimes q\in\mathbb{S}^{3}\label{eq:q_m_r}
\end{align}
In light of \eqref{Q2R}, the subtraction of two quaternions $q_{1},q_{2}\in\mathbb{S}^{3}$
is given by:
\begin{equation}
	q_{1}\boxminus q_{2}:=r_{q}(q_{1}\otimes q_{2}^{-1})\in\mathbb{R}^{3}\label{eq:q-q}
\end{equation}
Consider a set of quaternions $Q=\{q_{i}\in\mathbb{S}^{3}\}$ and
their corresponding weights $W=\{w_{i}\in\mathbb{R}\}$. To compute
the weighted average of these quaternions, the matrix $E$ is first
constructed as: 
\[
E=\sum w_{i}q_{i}q_{i}^{\top}\in\mathbb{R}^{4\times4}
\]
Next, the quaternion weighted mean $\text{QWM}(Q,W)$ is the eigenvector
corresponding to the largest magnitude eigenvalue of $E$ such that:
\begin{equation}
	\text{QWM}(Q,W)=\text{EigVector}(E)_{i}\in\mathbb{S}^{3}\label{eq:weighted_average}
\end{equation}
where $i=\text{argmax}(|\text{EigValue}(E)_{i}|)\in\mathbb{R}$. A
$d$-dimensional random variable (RV) $h\in\mathbb{R}^{d}$ drawn
from a Gaussian distribution with a mean $\overline{h}\in\mathbb{R}^{d}$
and a covariance matrix $C_{h}\in\mathbb{R}^{d\times d}$ is represented
by the following:
\[
h\sim\mathcal{N}(\overline{h},C_{h})
\]
Note that the expected value of $h$, denoted by $\mathbb{E}(h)$,
is equal to $\overline{h}$. The Gaussian (Normal) probability density
function of $h$ is formulated below:
\begin{align*}
	\mathbb{P}(h) & =\mathcal{N}(h|\overline{h},C_{h})\\
	& =\frac{\exp\left(-\frac{1}{2}(h-\overline{h})^{\top}C_{h}^{-1}(h-\overline{h})\right)}{\sqrt{(2\pi)^{d}\det(C_{h})}}\in\mathbb{R}
\end{align*}
where $\mathbb{P}(h)$ is the probability density of $h$.

\section{Problem Formulation\label{sec:Problem-Formulation}}

In this section, the kinetic and measurement models are introduced.
After defining the state vector, a state transition function is established
to define the relation between navigation state and the input data.
Moreover, the interdependence between the state and measurements vector
is formulated, which is essential for the proposed DeepUKF-VIN performance.

\subsection{Navigation Model in 3D}

The true navigation kinematics of a vehicle travelling in 3D space
are represented by \cite{hashim2021geometricNAV,hashim2021gps}: 
\begin{equation}
	\left\{ \begin{aligned}\dot{q} & =\frac{1}{2}\Gamma(\omega)q\in\mathbb{S}^{3}\\
		\dot{p} & =v\in\mathbb{R}^{3}\\
		\dot{v} & =g+R_{q}(q)a\in\mathbb{R}^{3}
	\end{aligned}
	\right.\label{c_dyn}
\end{equation}
with
\[
\Gamma(\omega)=\left[\begin{array}{cc}
	0 & -\omega^{\top}\\
	\omega & -[\omega]_{\times}
\end{array}\right]\in\mathbb{R}^{4\times4}
\]
where $q$ describe vehicle's orientation with respect to quaternion,
$\omega\in\mathbb{R}^{3}$ and $a\in\mathbb{R}^{3}$ denote angular
velocity and acceleration, respectively, while $p\in\mathbb{R}^{3}$
and $v\in\mathbb{R}^{3}$ refer to vehicle's position and linear velocity,
respectively, with $q,\omega,a\in\{\mathcal{B}\}$ and $p,v\in\{\mathcal{W}\}$.
In light of \cite{hashim2021geometricNAV}, the kinematics in \eqref{c_dyn}
is equivalent to:
\begin{equation}
	\left[\begin{array}{c}
		\dot{q}\\
		\dot{p}\\
		\dot{v}\\
		0
	\end{array}\right]=\underbrace{\left[\begin{array}{cccc}
			\frac{1}{2}\Gamma(\omega)q & 0 & 0 & 0\\
			0 & 0 & \mathbf{I}_{3} & 0\\
			0 & 0 & 0 & g+R_{q}(q)a\\
			0 & 0 & 0 & 0
		\end{array}\right]}_{M^{c}(q,\omega,a)}\left[\begin{array}{c}
		q\\
		p\\
		v\\
		1
	\end{array}\right]\label{eq:dyn_cont}
\end{equation}
Since the onboard data processor operating in discrete space and the
sensor data are updated at discrete instances, the continuous kinematics
in \eqref{eq:dyn_cont} need to be discretized. The true discrete
value at the $k$th time-step of $q\in\mathbb{S}^{3}$, $\omega\in\mathbb{R}^{3}$,
$a\in\mathbb{R}^{3}$, $p\in\mathbb{R}^{3}$, and $v\in\mathbb{R}^{3}$
is defined by $q_{k}\in\mathbb{S}^{3}$, $\omega_{k}\in\mathbb{R}^{3}$,
$a_{k}\in\mathbb{R}^{3}$ $p_{k}\in\mathbb{R}^{3}$, and $v_{k}\in\mathbb{R}^{3}$,
respectively. The equivalent discretized kinematics of the expression
in \eqref{eq:dyn_cont} is \cite{hashim2021geometricNAV}
\begin{equation}
	\left[\begin{array}{c}
		q_{k}\\
		p_{k}\\
		v_{k}\\
		1
	\end{array}\right]=\exp(M_{k-1}^{c}dT)\left[\begin{array}{c}
		q_{k-1}\\
		p_{k-1}\\
		v_{k-1}\\
		1
	\end{array}\right]\label{eq:dyn_dis}
\end{equation}
where $M_{k-1}^{c}=M^{c}(q_{k-1},\omega_{k-1},a_{k-1})$ and $dT$
denote a sample time.

\subsection{Measurement Model and Setup}

The IMU measurements at time step $k$ (angular velocity $\omega_{m,k}\in\mathbb{R}^{3}$
and acceleration $a_{m,k}\in\mathbb{R}^{3}$) and the related bias
in readings ($b_{\omega,k}$ and $b_{a,k}$) are as follows \cite{hashim2019Ito,hashim2020systematic}:
\begin{equation}
	\begin{cases}
		\text{IMU} & \left\{ \begin{aligned}\omega_{m,k} & =\omega_{k}+b_{\omega,k}+\eta_{\omega,k}\\
			a_{m,k} & =a_{k}+b_{a,k}+\eta_{a,k}
		\end{aligned}
		\right.\\
		\text{Bias} & \left\{ \begin{aligned}b_{\omega,k} & =b_{\omega,k-1}+\eta_{b\omega,k}\\
			b_{a,k} & =b_{a,k-1}+\eta_{ba,k}
		\end{aligned}
		\right.
	\end{cases}\label{a_m}
\end{equation}
where $\eta_{\omega,k}$ and $\eta_{a,k}$ refer to gyroscope and
accelerometer additive zero-mean white noise, respectively, while
the zero-mean white noise terms $\eta_{ba,k-1}$, and $\eta_{b\omega,k-1}\in\mathbb{R}^{3}$
correspond to $b_{a,k}$ and $b_{\omega,k}\in\mathbb{R}^{3}$. In
other words
\begin{equation}
	\left\{ \begin{aligned}\eta_{\omega,k} & \sim\mathcal{N}(0_{3},C_{\eta_{\omega,k}})\\
		\eta_{a,k} & \sim\mathcal{N}(0_{3},C_{\eta_{a,k}})\\
		\eta_{ba,k} & \sim\mathcal{N}(0_{3},C_{\eta_{b\omega,k}})\\
		\eta_{b\omega,k} & \sim\mathcal{N}(0_{3},C_{\eta_{ba,k}})
	\end{aligned}
	\right.\label{eq:noiseN}
\end{equation}
Note that if the noise vectors in \eqref{eq:noiseN} are assumed to
be uncorrelated, their covariance matrices will be diagonal with positive
entries, such that \cite{hashim2019Ito,hashim2020systematic}:
\begin{equation}
	\begin{cases}
		C_{\eta_{\omega,k}} & ={\rm diag}(c_{\eta_{\omega,k}}^{2})\\
		C_{\eta_{a,k}} & ={\rm diag}(c_{\eta_{a,k}}^{2})\\
		C_{\eta_{b\omega,k}} & ={\rm diag}(c_{\eta_{b\omega,k}}^{2})\\
		C_{\eta_{ba,k}} & ={\rm diag}(c_{\eta_{ba,k}}^{2})
	\end{cases}\label{eq:noise_elems}
\end{equation}
where $c_{\eta_{\omega,k}}$, $c_{\eta_{a,k}}$, $c_{\eta_{b\omega,k}}$,
and $c_{\eta_{ba,k}}\in\mathbb{R}^{3}$ represent the square roots
of the diagonal elements of their respective covariance matrices.
Let us define the state vector $x_{k}\in\mathbb{R}^{d_{x}}$ and the
augmented state vector $x_{k}^{a}\in\mathbb{R}^{d_{a}}$ such that
\begin{equation}
	\begin{cases}
		x_{k} & =\begin{bmatrix}q_{k}^{\top} & p_{k}^{\top} & v_{k}^{\top} & b_{\omega,k}^{\top} & b_{a,k}^{\top}\end{bmatrix}^{\top}\in\mathbb{R}^{d_{x}}\\[1em]
		x_{k}^{a} & =\begin{bmatrix}x_{k}^{\top} & \eta_{x,k}^{\top}\end{bmatrix}\in\mathbb{R}^{d_{a}}
	\end{cases}\label{eq:state}
\end{equation}
with $d_{x}=16$ and $d_{a}=22$ representing the dimensions of the
state vector and the augmented state vector, respectively, and $\eta_{x,k}$
representing the augmented noise vector, such that
\begin{equation}
	\eta_{x,k}=\begin{bmatrix}\eta_{\omega,k}^{\top} & \eta_{a,k}^{\top}\end{bmatrix}^{\top}\in\mathbb{R}^{d_{\eta_{x}}}\label{eq:eta_x}
\end{equation}
where $d_{\eta_{x}}=6$ is the dimension of the augmented noise. Consider
formulating the additive noise vector such that:
\begin{equation}
	\eta_{w,k}=\begin{bmatrix}0_{10}^{\top} & n_{b\omega,k}^{\top} & n_{ba,k}^{\top}\end{bmatrix}^{\top}\in\mathbb{R}^{d_{x}}\label{eq:add_noise}
\end{equation}
Then, the expression in \eqref{eq:dyn_dis}, using \eqref{a_m}, \eqref{eq:state},
\eqref{eq:eta_x}, and \eqref{eq:add_noise} can be written in form
of state transition function $\operatorname{f}:\mathbb{R}^{d_{a}}\rightarrow\mathbb{R}^{d_{x}}$
such that
\begin{equation}
	x_{k}=\operatorname{f}(x_{k-1}^{a},u_{k-1})+\eta_{w,k-1}\label{eq:state_transition}
\end{equation}
with the input vector being defined as $u_{k-1}=[\omega_{m,k-1}^{\top},a_{m,k-1}^{\top}]^{\top}\in\mathbb{R}^{d_{u}}$
and $d_{u}=6$ being the dimension of the input vector. Let the landmark
coordinates in $\{\mathcal{W}\}$ be represented as $\{l_{w,k,i}\in\mathbb{R}^{3}\}_{i}$,
where these coordinates are either known from prior information or
obtained from a series of stereo camera measurements. Similarly, let
the corresponding coordinates measured by the latest stereo camera
data in $\{\mathcal{B}\}$ be denoted as $\{l_{b,k,i}\in\mathbb{R}^{3}\}_{i}$,
where $i=\{1,\dots,d_{l,k}\}$ represents the index of each measured
landmark, and $d_{l,k}\in\mathbb{R}$ denotes the number of landmarks
at each measurement step. Note that $d_{l,k}$ is not constant and
may change at each step. We then construct the concatenated vectors
$l_{w,k}\in\mathbb{R}^{d_{z,k}}$, and $l_{b,k}\in\mathbb{R}^{d_{z,k}}$
such that:
\[
\begin{cases}
	l_{w,k} & =\left[l_{w,k,1}^{\top},\dots,l_{w,k,d_{l,k}}^{\top}\right]^{\top}\in\mathbb{R}^{d_{z,k}}\\
	l_{b,k} & =\left[l_{b,k,1}^{\top},\dots,l_{b,k,d_{l}}^{\top}\right]^{\top}\in\mathbb{R}^{d_{z,k}}
\end{cases}
\]
where $d_{z,k}=3d_{l,k}$ represents the dimension of the measurement
vector $z_{k}=l_{b},k$. The $i$th measurement function $\operatorname{h_{i}}:\mathbb{R}^{d_{x}}\times\mathbb{R}^{3}\rightarrow\mathbb{R}^{3}$
is defined as:
\begin{equation}
	\operatorname{h_{i}}(x_{k},l_{w,i})={R_{q}(q_{k})}^{\top}\left(l_{w,i}-p_{k}\right)\in\mathbb{R}^{3}\label{eq:hi}
\end{equation}
where $q_{k},p_{k}\subset x_{k}$ (see \eqref{eq:state}). The measurement
function $\operatorname{h}:\mathbb{R}^{d_{x}}\times\mathbb{R}^{d_{z,k}}\rightarrow\mathbb{R}^{d_{z,k}}$
is given by:
\begin{equation}
	\operatorname{h}(x_{k},l_{w})=\left[{\operatorname{h_{i}}(x_{k},l_{w,0})}^{\top},\dots,{\operatorname{h_{i}}(x_{k},l_{w,d_{l}})}^{\top}\right]^{\top}\in\mathbb{R}^{d_{z,k}}\label{eq:h}
\end{equation}
The measurement function in \eqref{eq:h} is used to find the measurement
vector $z_{k}$ at each time step $k$ such that
\begin{equation}
	z_{k}=\operatorname{h}(x_{k},l_{w})+\eta_{l,k}\in\mathbb{R}^{d_{z,k}}\label{eq:measurement_full}
\end{equation}
where $\eta_{l,k}\sim\mathcal{N}(0_{d_{z,k}},C_{\eta_{l},k})$ is
the measurement additive white noise. The covariance matrix $C_{\eta_{l},k}\in\mathbb{R}^{d_{z,k}\times d_{z,k}}$
is defined by: 
\begin{equation}
	C_{\eta_{l},k}=c_{\eta_{l},k}^{2}\mathbf{I}_{d_{z,k}}\label{eq:c_l}
\end{equation}
where $c_{\eta_{l},k}\in\mathbb{R}$ is a scalar. This definition
is particularly useful since $d_{z,k}$ may vary at each time step
$k$.

\section{Quaternion-based UKF-VIN\label{sec:QNUKF}}

This section provides a detailed description of the quaternion-based
Unscented Kalman Filter for 3D Visual-Inertial Navigation (UKF-VIN)
design which will be subsequently tightly-coupled with the proposed
Deep Learning-based Adaptation Mechanism (DLAM) for adaptive tuning
of UKF-VIN covariance matrices. The proposed approach builds upon
the standard UKF \cite{Khashayar2025_TIM_Hashim}, incorporating specific
modifications to address challenges inherent in navigation tasks.
These adaptations ensure the UKF operates effectively within the quaternion
space $\mathbb{S}^{3}$, preserving the physical validity of orientation
estimation. Furthermore, the design accommodates the intermittent
nature of vision data, which is not available at every time step,
while consistently integrating IMU data.

\subsection{Initialization}

he filter is initialized with the initial state vector estimate $\hat{x}_{0|0}\in\mathbb{R}^{d_{x}}$,
and its associated covariance estimate $P_{0|0}\in\mathbb{R}^{(d_{x}-1)\times(d_{x}-1)}$which
represents the confidence in the initial state estimate. The reduced
dimensionality of the covariance matrix arises from the fact that
the quaternion in the state vector has three degrees of freedom, despite
having four components \cite{mourikis2007multi}.

\subsection{Aggregate Predict\label{sec:batch_pred}}

At each time step $k$ where image data is available, the current
state vector is predicted using the last $d_{b}$ input vectors $u_{k-1-d_{b}:k-1}\in\mathbb{R}^{d_{b}\times d_{u}}$,
along with the previous state vector estimate $\hat{x}_{k-1-d_{b}|k-1}\in\mathbb{R}^{d_{x}}$
and the covariance matrix $P_{k-1-d_{b}|k-1}\in\mathbb{R}^{(d_{x}-1)\times(d_{x}-1)}$,
the current state vector is predicted. Here $d_{b}$ represents the
number of measurements received from the IMU between the current and
the last instance when image data was available. For each $j=\{k-1-d_{b},\dots,k-1\}$,
the following steps are executed sequentially.

\subsubsection{Augmentation\label{sec:augmentation}}

The augmented state vector $x_{j}^{a}\in\mathbb{R}^{d_{a}}$, and
the augmented covariance matrix $P_{j|j}^{a}\in\mathbb{R}^{(d_{a}-1)|(d_{a}-1)}$
are constructed as follows:
\begin{align}
	\hat{x}_{j|j}^{a} & =\left[\hat{x}_{j|j}^{\top},0_{m_{n_{x}}\times1}^{\top}\right]^{\top}\in\mathbb{R}^{m_{a}}\label{eq:QNUKF_augment1}\\
	P_{j|j}^{a} & ={\rm diag}(P_{j|j},C_{\eta_{x},k-1})\in\mathbb{R}^{(m_{a}-1)\times(m_{a}-1)}\label{eq:QNUKF_augment2}
\end{align}
where $C_{\eta_{x},k}={\rm diag}(C_{\eta_{w,k}},C_{\eta_{a,k}})\in\mathbb{R}^{\eta_{x}}$
is the covariance matrix of $\eta_{x,k}$ as defined in \eqref{eq:eta_x}. 

\subsubsection{Sigma Points Construction}

Using the augmented estimate of the state vector and its covariance,
while accounting for the reduced dimensionality of the quaternions
and applying the unscented transform \cite{Khashayar2025_TIM_Hashim}, the sigma points
representing the prior distribution are computed as follows:
\begin{equation}
	\left\{ \begin{aligned}\chi_{j|j,0}^{a} & =\hat{x}_{j|j}^{a}\in\mathbb{R}^{d_{a}}\\
		\chi_{j|j,\nu}^{a} & =\hat{x}_{j|j}^{a}\boxplus\delta\hat{x}_{\nu}^{a}\in\mathbb{R}^{d_{a}}\hspace{0.3cm}\nu=\{1,\dots,2(d_{a}-1)\}_{\nu}\\
		\chi_{j|j,\nu+m_{a}}^{a} & =\hat{x}_{j|j}^{a}\boxminus\delta\hat{x}_{\nu}^{a}\in\mathbb{R}^{d_{a}}
	\end{aligned}
	\right.\label{eq:Sigma_QNUKF}
\end{equation}
where $\delta\hat{x}_{j,\nu}^{a}=\left(\sqrt{(d_{a}-1+\lambda)P_{j|j}^{a}}\right)_{\nu}\in\mathbb{R}^{m_{a}-1}$,
with the subscript $\nu$ representing the $\nu$th column. The operators
$\boxplus$ and $\boxminus$ in \eqref{eq:Sigma_QNUKF} are defined
in accordance with \eqref{eq:q_p_r} and \eqref{eq:q_m_r}, such that
\begin{align}
	\hat{x}_{j|j}^{a}\boxplus\delta\hat{x}_{\nu}^{a} & =\begin{bmatrix}\hat{x}_{j|j,q}^{a}\boxplus\delta\hat{x}_{j,\nu,r}^{a}\\
		\hat{x}_{j|j,-}^{a}+\delta\hat{x}_{j,\nu,-}^{a}
	\end{bmatrix}\in\mathbb{R}^{d_{a}}\label{eq:x+dx}\\
	\hat{x}_{j|j}^{a}\boxminus\delta\hat{x}_{\nu}^{a} & =\begin{bmatrix}\hat{x}_{j|j,q}^{a}\boxminus\delta\hat{x}_{j,\nu,r}^{a}\\
		\hat{x}_{j|j,-}^{a}-\delta\hat{x}_{j,\nu,-}^{a}
	\end{bmatrix}\in\mathbb{R}^{d_{a}}\label{eq:x-dx}
\end{align}
where $\hat{x}_{j|j,q}^{a}\in\mathbb{S}^{3}$ and $\hat{x}_{j|j,-}^{a}\in\mathbb{R}^{d_{a}-4}$
represent the quaternion and non-quaternion components $\hat{x}_{j|j}^{a}\in\mathbb{R}^{d_{a}}$,
respectively, and $\delta\hat{x}_{j,\nu,r}^{a}\in\mathbb{R}^{3}$
and $\delta\hat{x}_{j,\nu,-}^{a}\in\mathbb{R}^{d_{a}-4}$ denote the
rotation vector and non-rotation vector components of $\delta\hat{x}_{j,\nu}^{a}\in\mathbb{R}^{d_{a}-1}$,
respectively. Note that $\lambda\in\mathbb{R}$ is a tuning parameter
that controls the spread of the sigma points. 

\subsubsection{Sigma Points Propagation}

In this step, the sigma points defined in \eqref{eq:Sigma_QNUKF}
are propagated through the state transition function \eqref{eq:state_transition}
to obtain the predicted sigma points $\{\chi_{j+1|j,\nu}\}_{\nu}$
such that
\begin{equation}
	\chi_{j+1|j,\nu}=\operatorname{f}(\chi_{j|j,\nu}^{a},u_{j})\in\mathbb{R}^{d_{x}}\hspace{0.5cm}\nu=\{1,\dots,2(d_{a}-1)\}_{\nu}\label{eq:propagate_sigma_points_augmented}
\end{equation}

\subsubsection{Calculate the Predicted Mean and Covariance}

The weighted mean $\hat{x}_{j+1|j}\in\mathbb{R}^{d_{x}}$ and covariance
$P_{j+1|j}\in\mathbb{R}^{(d_{x}-1)\times(d_{x}-1)}$ of the predicted
sigma points $\{\chi_{j+1|j,\nu}\}_{\nu}$ are determined in this
step in accordance with \eqref{eq:weighted_average}. These quantities
are calculated as follows:
\begin{align}
	\hat{x}_{j+1|j} & =\begin{bmatrix}\text{QWM}(\{\chi_{j+1|j,\nu,q}\}_{\nu},\{w_{\nu}^{m}\}_{\nu})\\
		{\displaystyle \sum_{\nu=0}^{2(d_{a}-1)}w_{\nu}^{m}\mathcal{X}_{j+1|j,\nu,-}}
	\end{bmatrix}\in\mathbb{R}^{d_{x}}\label{eq:state_mean_QNUKF}
\end{align}
\begin{align}
	P_{j+1|j} & =\sum_{\nu=0}^{2(d_{a}-1)}\left[w_{\nu}^{c}(\chi_{j+1|j,\nu}\boxminus\hat{x}_{j+1|j})(\chi_{j+1|j,\nu}\boxminus\hat{x}_{j+1|j})^{\top}\right]\nonumber \\
	& \qquad+C_{\eta_{w},k-1}\in\mathbb{R}^{(d_{x}-1)\times(d_{x}-1)}\label{eq:state_cov_QNUKF}
\end{align}
with $\chi_{j+1|j,\nu,q}\in\mathbb{S}^{3}$ and $\chi_{j+1|j,\nu,-}\in\mathbb{R}^{d_{x}-4}$
representing the quaternion and non-quaternion components of $\chi_{j+1|j,\nu}\in\mathbb{R}^{d_{x}}$,
respectively. Note that in \eqref{eq:state_mean_QNUKF}, the quaternion
weighted average \eqref{eq:weighted_average} is used for quaternion
components of the propagated sigma point vectors and the straightforward
weighted average for non-orientation components of the propagated
sigma point vector. The weights $\{w_{\nu}^{m}\}_{\nu}$ and $\{w_{\nu}^{c}\}_{\nu}$
in \eqref{eq:state_mean_QNUKF} and \eqref{eq:state_cov_QNUKF} are
derived from:
\begin{equation}
	\left\{ \begin{aligned}w_{0}^{m} & =\frac{\lambda}{\lambda+(d_{a}-1)}\in\mathbb{R}\\
		w_{0}^{c} & =\frac{\lambda}{\lambda+(d_{a}-1)}+1-\alpha^{2}+\beta\in\mathbb{R}\\
		w_{\nu}^{m} & =w_{\nu}^{c}=\frac{1}{2((d_{a}-1)+\lambda)}\in\mathbb{R}\\
		& \hspace{2cm}\nu=\{1,\ldots,2(d_{a}-1)\}_{\nu}
	\end{aligned}
	\right.\label{eq:weights_QNUKF}
\end{equation}
where $\alpha$ and $\beta\in\mathbb{R}$ are tuning parameters. The
$\boxminus$ operator in \eqref{eq:state_cov_QNUKF} is defined in
accordance with \eqref{eq:q-q} as follows:
\begin{align}
	\chi_{j+1|j,\nu}\boxminus\hat{x}_{j+1|j} & =\begin{bmatrix}\chi_{j+1|j,\nu,q}\ominus\hat{x}_{j+1|j,q}\\
		\chi_{j+1|j,\nu,-}-\hat{x}_{j+1|j,-}
	\end{bmatrix}\in\mathbb{R}^{d_{x}-1}\label{eq:x_x}
\end{align}
where $\hat{x}_{j+1|j,q}\in\mathbb{S}^{3}$, and $\hat{x}_{j+1|j,-}\in\mathbb{R}^{d_{x}-4}$
represent the quaternion and non-quaternion components of $\hat{x}_{j+1|j}\in\mathbb{R}^{d_{x}}$,
respectively. Note that $C_{\eta_{w},k}={\rm diag}(0_{d_{x}-7},C_{\eta_{w,k}},C_{\eta_{a,k}})\in\mathbb{R}^{(d_{x}-1)\times(d_{x}-1)}$
denotes the covariance matrix of $\eta_{w,k}$ as defined in \eqref{eq:add_noise}.

\subsubsection{Iterate over batch}

If $j=k-1$, corresponding to the end of the batch, the predicted
state estimate $\hat{x}_{k|k-1}$, the covariance $P_{k|k-1}$, and
the predicted sigma points $\{\chi_{j+1|j,\nu}\}_{\nu}$, as defined
in \eqref{eq:state_mean_QNUKF}, \eqref{eq:state_cov_QNUKF}, and
\eqref{eq:propagate_sigma_points_augmented}, respectively, are passed
to the update step (see Section \ref{sec:update}). Otherwise, $\hat{x}_{j+1|j+1}$
and $P_{j+1|j+1}$ are set to $\hat{x}_{j+1|j}$ and $P_{j+1|j}$,
respectively. The aggregate prediction algorithm then increments $j\leftarrow j+1$
and continues by returning to Section \ref{sec:augmentation}.

\subsection{Update\label{sec:update}}

\subsubsection{Calculate Measurement Sigma Points And Its Statistics}

At time step $k$, the predicted sigma points $\{\chi_{j+1|j,\nu}\}_{\nu}$
are passed through the measurement function \eqref{eq:h} to calculate
the measurement sigma points $\left\{ \zeta_{k,\nu}\in\mathbb{R}^{d_{z,k}}\right\} _{\nu}$
such that
\begin{align}
	\zeta_{k,\nu} & =h(\chi_{k|k-1,\nu},l_{w})\in\mathbb{R}^{d_{z,k}}\label{eq:propagate_measurment}
\end{align}
Considering \eqref{eq:measurement_full} and \eqref{eq:weights_QNUKF},
the expected value and covariance of $\left\{ \zeta_{k,\nu}\right\} _{\nu}$,
denoted by $\hat{z}_{k}\in\mathbb{R}^{d_{z,k}}$ and $P_{z_{k}}\in\mathbb{R}^{d_{z,k}\times d_{z,k}}$,
respectively, are determined as follows:
\begin{align}
	\hat{z}_{k} & =\sum_{\nu=0}^{2(d_{a}-1)}w_{\nu}^{m}\zeta_{k,\nu}\label{eq:zhat_QNUKF}\\
	P_{z_{k}} & =\sum_{\nu=0}^{2(d_{a}-1)}w_{\nu}^{c}[\zeta_{k,\nu}-\hat{z}_{k}][\zeta_{k,\nu}-\hat{z}_{k}]^{\top}\nonumber \\
	& \hspace{1.5cm}+C_{\eta_{l},k}\label{eq:P_zz_QUKF}
\end{align}
Using \eqref{eq:weights_QNUKF}, \eqref{eq:x_x}, and \eqref{eq:zhat_QNUKF},
the cross covariance matrix $P_{x_{k},z_{k}}\in\mathbb{R}^{(d_{x}-1)\times d_{z,k}}$
is calculated by
\begin{align}
	P_{x_{k},z_{k}} & =\sum_{\nu=0}^{2(d_{a}-1)}w_{j}^{c}[\chi_{k|k-1,\nu}\boxminus\hat{x}_{k|k-1}][\zeta_{k,\nu}-\hat{z}_{k}]^{\top}\label{eq:P_xz_QUKF}
\end{align}
Note that the operator $\boxminus$ in \eqref{eq:P_xz_QUKF} is defined
in \eqref{eq:x_x}. 

\subsubsection{Calculate The Current State Estimate}

First, the Kalman gain $K_{k}\in\mathbb{R}^{(d_{x}-1)\times d_{z,k}}$,
based on \eqref{eq:P_xz_QUKF} and \eqref{eq:P_zz_QUKF}, is computed
as 
\begin{equation}
	K_{k}=P_{x_{k},z_{k}}P_{z_{k},z_{k}}^{\top}\in\mathbb{R}^{(d_{x}-1)\times d_{z,k}}\label{eq:K}
\end{equation}
The correction vector $\delta\hat{x}_{k}\in\mathbb{R}^{d_{x}-1}$
is then derived, using \eqref{eq:zhat_QNUKF} and \eqref{eq:K}, as
\begin{equation}
	\delta\hat{x}_{k}=K_{k}(z_{k}-\hat{z}_{k})\in\mathbb{R}^{d_{x}-1}\label{eq:correction vector}
\end{equation}
Representing the rotation and non-rotation components of $\delta\hat{x}_{k}\in\mathbb{R}^{d_{x}-1}$
as $\delta\hat{x}_{k,r}\in\mathbb{S}^{3}$ and $\delta\hat{x}_{k,-}\in\mathbb{R}^{d_{x}-4}$,
respectively, the updated state estimate $\hat{x}_{k|k}$ is computed
as 
\begin{equation}
	\hat{x}_{k|k}=\hat{x}_{k|k-1}\boxplus\delta\hat{x}_{k}\label{eq:final_updateQUNKF}
\end{equation}
Note that the $\boxplus$ operator in \eqref{eq:final_updateQUNKF}
has been defined in \eqref{eq:x+dx}. Finally, the covariance matrix
associated with this state estimate is updated, based on \eqref{eq:state_cov_QNUKF},
\eqref{eq:P_zz_QUKF}, and \eqref{eq:K}, as 
\begin{equation}
	P_{k|k}=P_{k|k-1}-K_{k}P_{z_{k},z_{k}}K_{k}^{\top}\label{eq:update:P}
\end{equation}

\subsection{Iterate and Collect IMU Measurements}

Proceed to the next index by setting $k\leftarrow k+1$. The IMU measurements
$u_{k-1-d_{b}:k-1}\in\mathbb{R}^{d_{b}\times d_{u}}$ are collected
until image data becomes available at $z_{k}$, at which point the
algorithm proceeds to \ref{sec:batch_pred}. If image data is not
yet available, the collection of IMU data continues.

\section{Deep Learning-based Adaptation Mechanism (DLAM)\label{sec:DLAM}}

This section provides a detailed discussion of the design of the proposed
DLAM. This mechanism adaptively updates the covariance matrices of
the noise parameters used by the quaternion-based UKF-VIN at each
time step, leveraging the input data. The DLAM is designed to enhance
the filter's performance by dynamically adjusting to changes in noise
characteristics, thereby ensuring more accurate and robust state estimation.
The proposed DLAM is composed of two neural networks, namely, the
IMU-Net (Section \ref{sec:IMU-Net}) and the Vision-Net (Section \ref{sec:Vision-Net}).
The IMU-Net processes the last $d_{\text{GRU}}\in\mathbb{R}$ measurement
vectors (see \eqref{eq:state_transition}) as input, while the Vision-Net
takes the current stereo image measurements as input. Each network
produces scaling factors corresponding to their respective sensor
covariance matrices. Given that the IMU noise model includes 12 unknown
terms (see \eqref{eq:noise_elems}), and the vision unit covariance
matrix contains a single unknown element (see \eqref{eq:c_l}), the
IMU-Net and Vision-Net generate outputs of size 12 and 1, respectively.
This can be formulated as: 
\begin{align}
	\{\gamma_{i,k}\}_{i=\{1,\dots,12\}} & =\text{IMUNet}(u_{k-1-d_{\text{GRU}}:k-1},W_{IN}),\label{eq:IN}\\
	\gamma_{13,k} & =\text{VisionNet}(\text{img}_{l},\text{img}_{r},W_{VN}),\label{eq:VN}
\end{align}
where $u_{k-11:k-1}\in\mathbb{R}^{d_{\text{GRU}}\times d_{u}}$ represents
the last $d_{\text{GRU}}$ measurement vectors, $\text{img}_{l}$
and $\text{img}_{r}$ denote the left and right images, respectively,
and $W_{IN}$ and $W_{VN}$ represent the weights and biases of the
IMU-Net and Vision-Net, respectively. To explain the use of the scaling
parameters $\{\gamma_{i,k}\}_{i=\{1,\dots,13\}}$ generated by IMU-Net
and Vision-Net, let us define the standard deviation vector $c_{k}\in\mathbb{R}^{13}$
using the square root of the diagonal elements of the covariance matrices
in \eqref{eq:noise_elems} and \eqref{eq:c_l}, such that: 
\begin{equation}
	c_{k}=\begin{bmatrix}c_{\eta_{\omega,k}}^{\top} & c_{\eta_{a,k}}^{\top} & c_{\eta_{b\omega,k}}^{\top} & c_{\eta_{ba,k}}^{\top} & c_{\eta_{l},k}^{\top}\end{bmatrix}^{\top}\in\mathbb{R}^{13}\label{eq:c_k}
\end{equation}
For each $i$th element of $c_{k}$, denoted by $c_{k,i}\in\mathbb{R}$,
let $\bar{c}_{k,i}\in\mathbb{R}$ represent its nominal value, obtained
through traditional offline tuning methods. Then, at each time step
$k$, using the scaling parameters $\{\gamma_{i,k}\}_{i=\{1,\dots,13\}}$
from \eqref{eq:VN} and \eqref{eq:IN}, the standard deviation vector
elements defined in \eqref{eq:c_k} are computed as in \cite{Zhou2022IMUAlgorithm,brossard2020ai}:
\begin{equation}
	c_{k,i}=\bar{c}_{k,i}10^{\upsilon\tanh\gamma_{i,k}}\hspace{1cm}i=\{1,2,\dots,13\}\label{eq:c_k adap}
\end{equation}
where $\upsilon\in\mathbb{R}$ specifies the degree to which the predicted
$c_{k,i}$ may deviate from the nominal value $\bar{c}_{k,i}$. Considering
\eqref{eq:c_k adap}, \eqref{eq:c_k}, \eqref{eq:noise_elems}, and
\eqref{eq:c_l}, the covariance matrices used by the filter are found
as follows:
\begin{equation}
	\begin{cases}
		C_{\eta_{\omega,k}} & ={\rm diag}(c_{k,1:3}^{2})\\
		C_{\eta_{a,k}} & ={\rm diag}(c_{k,4:6}^{2})\\
		C_{\eta_{b\omega,k}} & ={\rm diag}(c_{k,7:9}^{2})\\
		C_{\eta_{ba,k}} & ={\rm diag}(c_{k,10:12}^{2})\\
		C_{\eta_{l,k}} & =c_{k,13}^{2}\mathbf{I}_{d_{z,k}}
	\end{cases}\label{eq:noise elems final}
\end{equation}
where $c_{k,i:j}\in\mathbb{R}^{j-i+1}$ denotes the $i$th to $j$th
components of $c_{k}$. 

\begin{figure}[h]
	\centering{}\includegraphics[width=1\columnwidth]{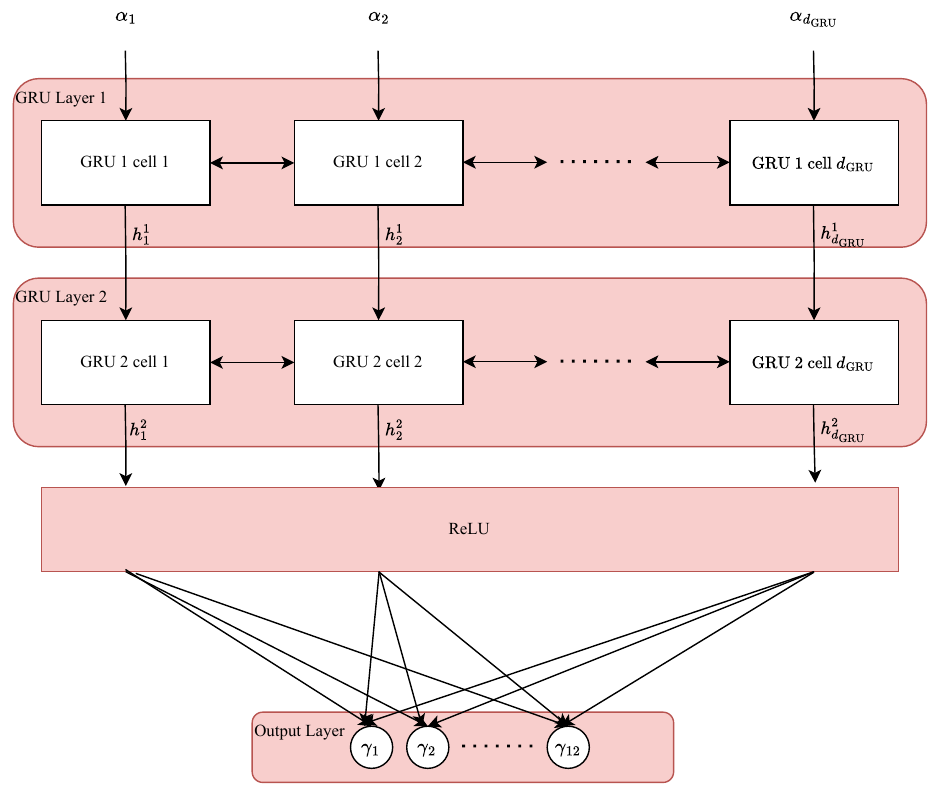} \caption{IMU-Net Architecture Schematics}
	\label{fig:IMU_Net}
\end{figure}

\subsection{IMU-Net\label{sec:IMU-Net}}

It is assumed that the covariance matrices $C_{\eta_{w,k}}$ and $C_{\eta_{a,k}}$
can be optimized for each batch by considering the input vector, which
comprises the last $d_{b}$ IMU measurements and practically it is
a feasible and realizable condition. Recurrent deep learning frameworks,
particularly Recurrent Neural Networks (RNNs) and their advanced variants,
have proven effective in modeling sequential data due to their capacity
to capture dependencies across time steps \cite{jernite2016variable}.
While traditional RNNs are foundational, they often struggle with
long-term dependencies due to challenges such as vanishing gradients
\cite{rusch2021unicornn,Lechner2020LearningLD}, leading to the development
of more sophisticated architectures, such as Long Short-Term Memory
(LSTM) networks \cite{hochreiter1997lstm} and Gated Recurrent Units
(GRUs) \cite{Dey2017GatevariantsOG}. LSTMs and GRUs were specifically
designed to address the limitations of standard RNNs by incorporating
gating mechanisms that regulate information flow, enabling more stable
long-term memory retention. In particular, GRUs offer a streamlined
architecture by combining the forget and input gates of LSTMs into
a single update gate, making them computationally more efficient while
retaining the ability to model complex temporal relationships. GRUs
have been shown to outperform LSTMs, particularly when the dataset
is small, while also being less computationally intensive \cite{Dey2017GatevariantsOG,Shewalkar2019PerformanceEO}.\vspace{0.3cm}

For a GRU cell at time step $l$, let $\alpha_{l}\in\mathbb{R}^{d_{u}}$
denote the GRU cell input vector. Each GRU cell computes its hidden
state $\overrightarrow{h}_{l}\in\mathbb{R}^{d_{h}}$ by leveraging
three key components: the update gate $\overrightarrow{z}_{l}\in\mathbb{R}^{d_{h}}$,
reset gate $\overrightarrow{r}_{l}\in\mathbb{R}^{d_{h}}$, and candidate
hidden state $\overrightarrow{n}_{l}\in\mathbb{R}^{d_{h}}$, where
$d_{h}$ represents the dimensionality of the hidden state. The equations
governing these components are as follows:
\vspace{0.3cm}
\begin{itemize}
	\item \textbf{Update Gate}:
	\begin{equation}
		\overrightarrow{z}_{l}=\sigma(\overrightarrow{W}_{z}\alpha_{l}+\overrightarrow{U}_{z}\overrightarrow{h}_{l-1}+\overrightarrow{b}_{z})\in\mathbb{R}^{d_{h}}\label{eq:gru_z}
	\end{equation}
	where $\overrightarrow{W}_{z}\in\mathbb{R}^{d_{h}\times d_{u}}$ and
	$\overrightarrow{U}_{z}\in\mathbb{R}^{d_{h}\times d_{h}}$ are weight
	matrices, and $\overrightarrow{b}_{z}\in\mathbb{R}^{d_{h}}$ is the
	bias term. Note that the function $\sigma:\mathbb{R}^{d_{h}}\rightarrow\mathbb{R}^{d_{h}}$
	denotes the sigmoid function. The update gate controls the degree
	to which the previous hidden state $\overrightarrow{h}_{l-1}$ is
	retained.\vspace{0.3cm}
	\item \textbf{Reset Gate}: 
	\begin{equation}
		\overrightarrow{r}_{l}=\sigma(\overrightarrow{W}_{r}\alpha_{l}+\overrightarrow{U}_{r}\overrightarrow{h}_{l-1}+\overrightarrow{b}_{r})\in\mathbb{R}^{d_{h}}\label{eq:gru_r}
	\end{equation}
	where $\overrightarrow{W}_{r}\in\mathbb{R}^{d_{h}\times d_{u}}$,
	$\overrightarrow{U}_{r}\in\mathbb{R}^{d_{h}\times d_{h}}$, and $\overrightarrow{b}_{r}\in\mathbb{R}^{d_{h}}$
	are the corresponding parameters for the reset gate, which determines
	the relevance of the previous hidden state in computing the candidate
	hidden state.\vspace{0.3cm}
	\item \textbf{Candidate Activation}: 
	\begin{equation}
		\overrightarrow{n}_{l}=\tanh(\overrightarrow{W}_{n}\alpha_{l}+\overrightarrow{r}_{l}\circ(\overrightarrow{U}_{n}\overrightarrow{h}_{l-1})+\overrightarrow{b}_{n})\label{eq:gru_ht}
	\end{equation}
	where the $\circ$ operator represents element-wise multiplication,
	and $\overrightarrow{W}_{n}\in\mathbb{R}^{d_{h}\times d_{u}}$, $\overrightarrow{U}_{n}\in\mathbb{R}^{d_{h}\times d_{h}}$,
	and $\overrightarrow{b}_{n}\in\mathbb{R}^{d_{h}}$ are the weight
	matrices and bias vector associated with the candidate hidden state.
	Note that $\tanh:\mathbb{R}^{d_{h}}\rightarrow\mathbb{R}^{d_{h}}$
	denotes the hyperbolic tangent function.\vspace{0.3cm}
	\item \textbf{Hidden State Update}: 
	\begin{equation}
		\overrightarrow{h}_{l}=\overrightarrow{z}_{l}\circ\overrightarrow{h}_{l-1}+(1-\overrightarrow{z}_{l})\circ\overrightarrow{n}_{l}\label{eq:gru_h}
	\end{equation}
	This update equation combines the previous hidden state $\overrightarrow{h}_{l-1}$
	and the candidate $\overrightarrow{n}_{l}$, governed by the update
	gate $\overrightarrow{z}_{l}$. 
\end{itemize}
\begin{figure*}[h]
	\centering \begin{tikzpicture}
		
		\node[canvas is zy plane at x=0] (left_input) at (0,0,0) {\includegraphics[width=3.8cm]{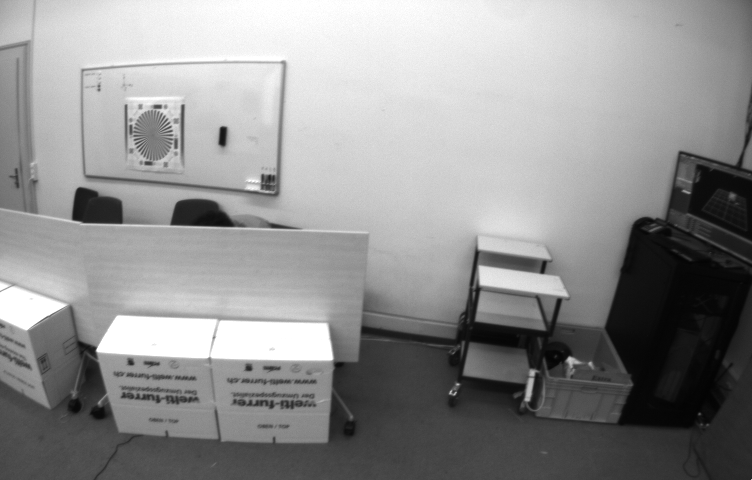}};
		\node[below=1.0cm of left_input] {\textbf{Left Image}};
		\node[canvas is zy plane at x=0] (right_input) at (0,-5,0) {\includegraphics[width=3.8cm]{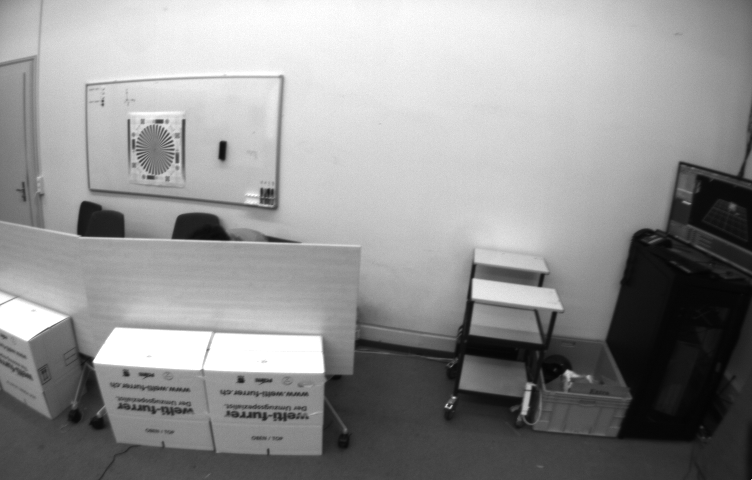}}; 
		\node[below=1.0cm of right_input] {\textbf{Right Image}};  
		
		\pic at (2,0,0) 
		{RightBandedBox={
				name=conv1, 
				caption=Conv1+ReLU $\mathbf{(16\times5\times5)}$, 
				width=2, 
				height=12, 
				depth=13, 
				fill=blue!20
		}};
		
		\pic at (2,-5,0) 
		{RightBandedBox={
				name=conv1_r, 
				caption=Conv1+ReLU $\mathbf{(16\times5\times5)}$, 
				width=2, 
				height=12, 
				depth=13, 
				fill=blue!20
		}};
		
		\pic at (4,0,0)
		{RightBandedBox={
				name=maxpool1, 
				caption=MaxPool $\mathbf{(4\times4)}$, 
				width=1.5, 
				height=9, 
				depth=9, 
				fill=purple!20
		}};
		
		\pic at (4,-5,0)
		{RightBandedBox={
				name=maxpool1_r, 
				caption=MaxPool $\mathbf{(4\times4)}$, 
				width=1.5, 
				height=9, 
				depth=9, 
				fill=purple!20
		}};
		
		\pic at (6,0,0) 
		{RightBandedBox={
				name=conv2, 
				caption=Conv2+ReLU $\mathbf{(32\times5\times5)}$, 
				width=2, 
				height=9, 
				depth=9, 
				fill=blue!30
		}};
		
		\pic at (6,-5,0) 
		{RightBandedBox={
				name=conv2_r, 
				caption=Conv2+ReLU $\mathbf{(32\times5\times5)}$, 
				width=2, 
				height=9, 
				depth=9, 
				fill=blue!30
		}};
		
		\pic at (8,0,0)
		{RightBandedBox={
				name=maxpool2, 
				caption=MaxPool $\mathbf{(4\times4)}$, 
				width=1.5, 
				height=6, 
				depth=6, 
				fill=purple!20
		}};
		
		\pic at (8,-5,0)
		{RightBandedBox={
				name=maxpool2_r, 
				caption=MaxPool $\mathbf{(4\times4)}$, 
				width=1.5, 
				height=6, 
				depth=6, 
				fill=purple!20
		}};
		
		\pic at (10,-2.5,0)  
		{RightBandedBox={
				name=flatten_concat, 
				caption=Flatten, 
				width=1, 
				height=15, 
				depth=1, 
				fill=yellow!20
		}};
		
		\pic at (12,-2.5,0)
		{RightBandedBox={
				name=fc_hid, 
				caption=Hidden $(32)$, 
				width=1, 
				height=4, 
				depth=1, 
				fill=cyan!20
		}};

		\pic at (14,-2.5,0)
		{RightBandedBox={
				name=fc_out, 
				caption=Output $\gamma_{13}$ $(1)$, 
				width=1, 
				height=1, 
				depth=1, 
				fill=magenta!20
		}};
		
		\draw[->, thick] (left_input) -- (conv1-west);
		\draw[->, thick] (right_input) -- (conv1_r-west);
		
		\draw[->, thick] (conv1-east) -- (maxpool1-west);
		\draw[->, thick] (conv1_r-east) -- (maxpool1_r-west);
		
		\draw[->, thick] (maxpool1-east) -- (conv2-west);
		\draw[->, thick] (maxpool1_r-east) -- (conv2_r-west);
		
		\draw[->, thick] (conv2-east) -- (maxpool2-west);
		\draw[->, thick] (conv2_r-east) -- (maxpool2_r-west);
		
		\draw[->, thick] (maxpool2-east) -- (flatten_concat-west);
		\draw[->, thick] (maxpool2_r-east) -- (flatten_concat-west);
		
		\draw[->, thick] (flatten_concat-east) -- (fc_hid-west);
		\draw[->, thick] (fc_hid-east) -- (fc_out-west);

	\end{tikzpicture} \caption{Vision-Net Architecture Schematics}
	\label{fig:IMU-Net}
\end{figure*}
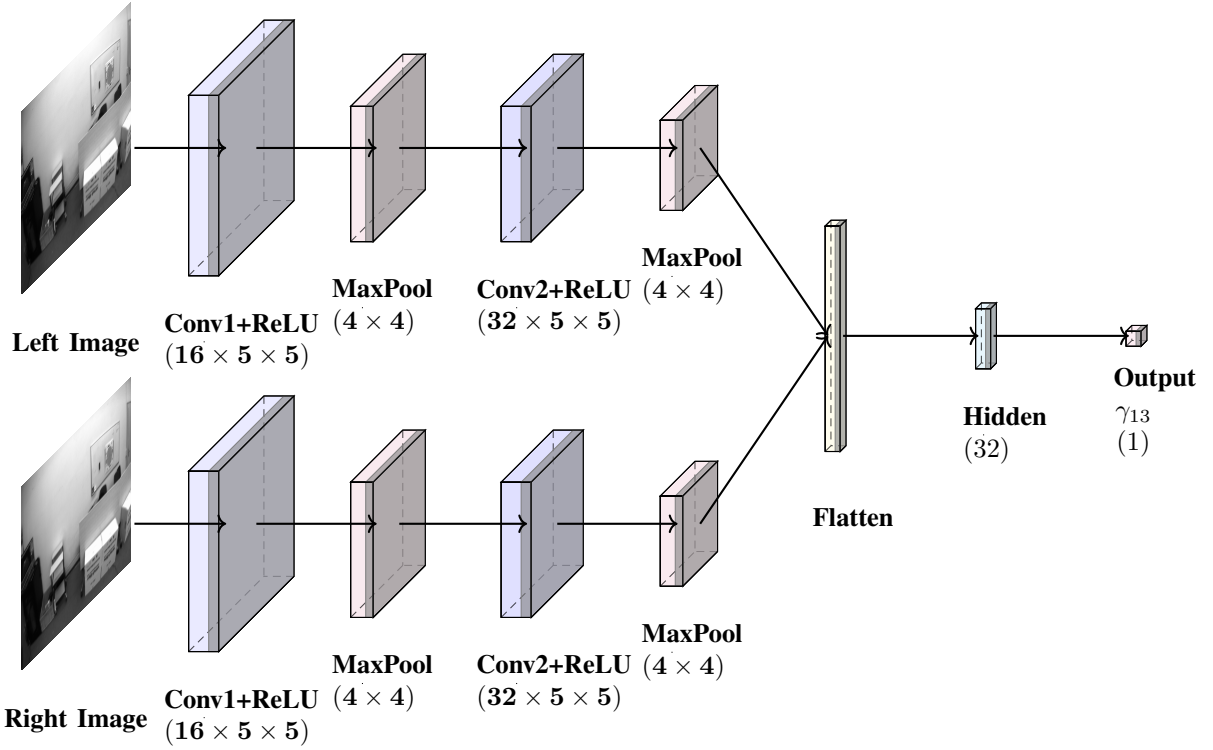
\begin{figure*}[!t]
	\vspace{-0.4cm}
	\centering \includegraphics[width=0.9\textwidth, height=0.41\textheight]{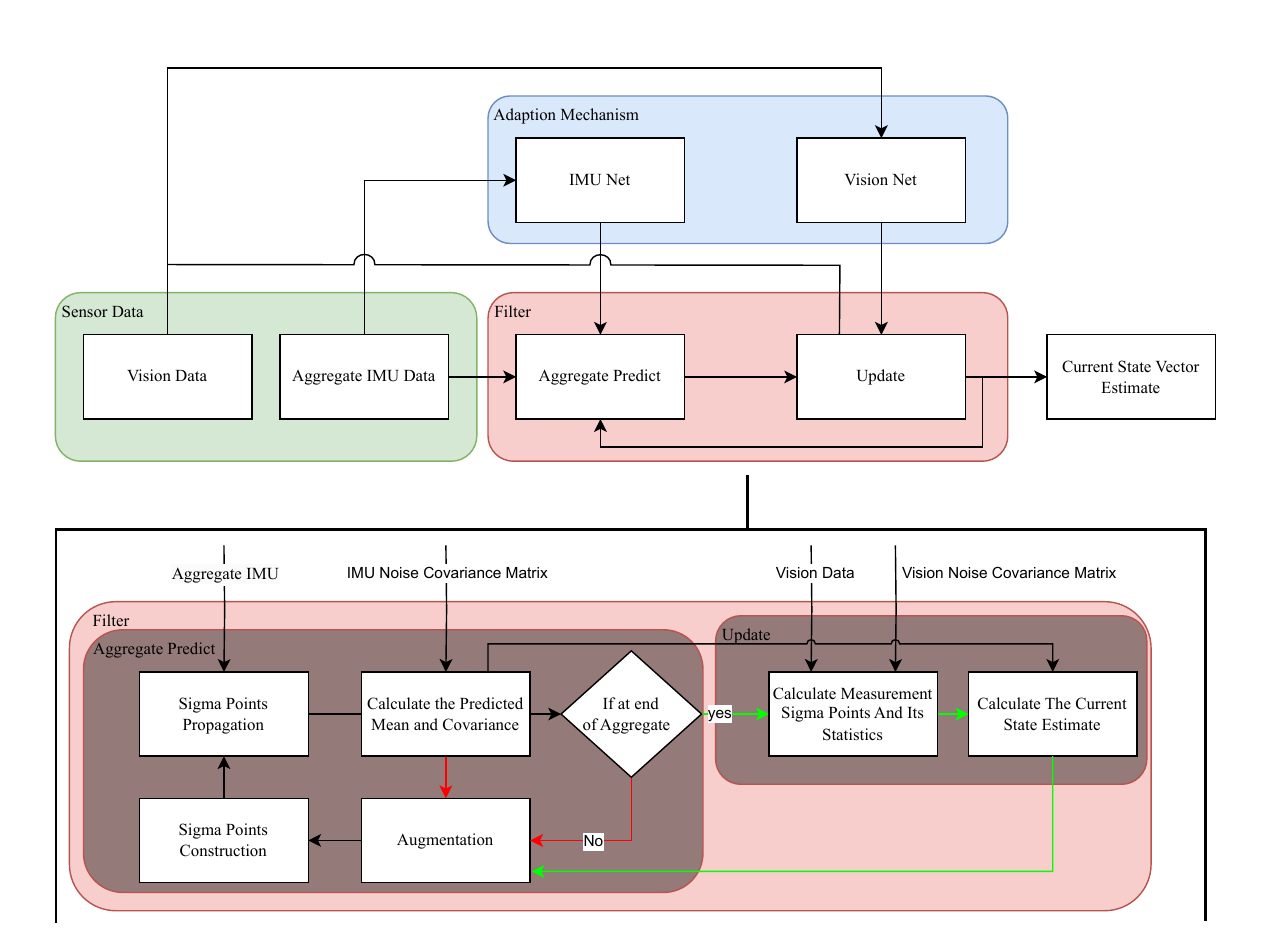}
	\caption{Summary schematic architecture of quaternion-based DeepUKF-VIN. First,
		the Aggregate Predict step of the filter is executed, incorporating
		the last known state information, aggregated IMU data, and the IMU
		noise covariance computed by IMU-Net. Next, the Update step is performed
		using the predicted state information and the vision covariance matrix
		estimated by Vision-Net. Raw IMU and vision data are used as inputs
		to IMU-Net and Vision-Net, respectively.}
	\label{fig:flowchart}
\end{figure*}
Equations \eqref{eq:gru_z}, \eqref{eq:gru_r}, \eqref{eq:gru_ht},
and \eqref{eq:gru_h} can be adapted to calculate the backward hidden
vector $\overleftarrow{h}_{l}$ by moving in reverse across the sequence,
computing each hidden state based on the subsequent hidden vector
$\overleftarrow{h}_{l+1}$ and the input vector $\alpha_{l}$. For
each GRU layer, consisting of $d_{\text{GRU}}$ GRU cells, both forward
and backward passes are computed. The forward and backward hidden
vectors for each cell are concatenated to form: 
\begin{equation}
	h_{l}=\begin{bmatrix}\overrightarrow{h}_{l}^{\top} & \overleftarrow{h}_{l}^{\top}\end{bmatrix}^{\top}\in\mathbb{R}^{2d_{h}}\label{eq:gruhconca}
\end{equation}
In summary, equations \eqref{eq:gru_z}, \eqref{eq:gru_r}, \eqref{eq:gru_ht},
\eqref{eq:gru_h}, and \eqref{eq:gruhconca} collectively define the
GRU function $\text{GRU}:\mathbb{R}^{d_{u}}\times\mathbb{R}^{d_{h}}\rightarrow\mathbb{R}^{2d_{h}}$
as follows: 
\[
h_{l}=\text{GRU}(\alpha_{l},\overrightarrow{h}_{l-1},\overleftarrow{h}_{l+1})
\]
To design IMU-Net, two layers of GRUs have been stacked with a fully
connected network at the end, with Rectified Linear Unit (ReLU) activation
function as the activation function between the GRUs and the fully
connected network, producing the scaling parameters (see \eqref{eq:IN}).
Note that the input to the first GRU layer (that is $\alpha_{1},\alpha_{2},\dots,\alpha_{d_{\text{GRU}}}$)
is set to the last $d_{\text{GRU}}$ IMU measurements ($u_{k-1-d_{\text{GRU}}:k-1}\in\mathbb{R}^{d_{\text{GRU}}\times d_{u}}$).
This process is visualized in Fig. \ref{fig:IMU_Net}.

\subsection{Vision-Net\label{sec:Vision-Net}}

The Vision-Net network is designed to adaptively estimate the measurement
covariance matrix based on vision data. The uncertainty in image measurements
can be effectively estimated from the current stereo-vision measurements.
In Vision-Net, each image is processed through a 2D convolutional
layer, followed by 2D max pooling, then a second 2D convolutional
layer, and another 2D max pooling layer. The resulting features are
flattened, concatenated, and subsequently passed through two fully
connected layers, ultimately producing the scaling parameter $\gamma_{13}\in\mathbb{R}$
(see \eqref{eq:VN}). Each convolutional layer is followed by a ReLU
activation function. A visualization of Vision-Net is provided in
Fig. \ref{fig:IMU-Net}.

\section{DeepUKF-VIN Training and Implementation\label{sec:Train}}

In summary, the DLAM-equipped UKF-VIN algorithm  referred to quaternion-based
DeepUKF-VIN, is illustrated in Fig. \ref{fig:flowchart}. For IMU-Net, we selected the last $d_{\text{GRU}}=10$ IMU measurements
as input (see \eqref{eq:IN}). This choice is based on the IMU's 200
Hz sample rate, compared to the image data's 20 Hz sample rate, meaning
there are at least 10 IMU measurements between each pair of vision
measurements. Although the structure of IMU-Net allows for variable-length
time series input data, using a fixed length of 10 measurements enhances
consistency and predictability. To train the models, a loss function
must be defined. Let the estimated orientation, position, and velocity
at time step $k$ be denoted by $\hat{q}_{k}$, $\hat{p}_{k}$, and
$\hat{v}_{k}$, respectively, with their corresponding estimation
errors denoted by $r_{e,k}$, $p_{e,k}$, and $v_{e,k}$. These errors
are defined as follows: 
\begin{equation}
	\begin{cases}
		r_{e,k} & =q_{k}\boxminus\hat{q}_{k}\in\mathbb{S}^{3}\\
		p_{e,k} & =p_{k}-\hat{p}_{k}\in\mathbb{R}^{3}\\
		v_{e,k} & =v_{k}-\hat{v}_{k}\in\mathbb{R}^{3}
	\end{cases}\label{eq:estimation_errors}
\end{equation}
For a set of $d^{\text{mini-batch}}\in\mathbb{R}$ estimations in
the $i$-th mini-batch, the total loss is computed as the weighted
sum of the mean square errors (MSE) of the individual errors defined
in \eqref{eq:estimation_errors}. The loss function for the mini-batch
is given by: 
\begin{equation}
	\begin{aligned}\text{Loss}_{i}^{\text{mini-batch}}= & w_{q}\frac{\sum\|r_{e,k}\|^{2}}{d^{\text{mini-batch}}}+w_{p}\frac{\sum\|p_{e,k}\|^{2}}{d^{\text{mini-batch}}}\\
		& +w_{v}\frac{\sum\|v_{e,k}\|^{2}}{d^{\text{mini-batch}}}
	\end{aligned}
	\label{eq:loss_batch}
\end{equation}
where $w_{q}$, $w_{p}$, and $w_{v}\in\mathbb{R}$ are the weights
that determine the relative importance of each term and are tuned
offline. As the steady-state performance of the filter is of greater
importance than its transient performance, the loss function in \eqref{eq:loss_batch}
is evaluated starting from the 51st data point onward. This ensures
that the loss function disregards the first 50 data points, which
represent the transient response of the filter.

The V1\_02\_medium part of the EuRoC dataset \cite{Burri25012016}
has been utilized for training. This dataset includes IMU measurements,
recorded at 200 Hz using the ADIS16448 sensor, mounted on an MAV.
Stereo images are captured as well by the Aptina MT9V034 global shutter
camera at a rate of 20 Hz. Ground truth data is provided at 200 Hz,
measured via the Vicon motion capture system. At each epoch, the whole
dataset will be utilized as a single batch. In other words, given
the initial state estimate and covariance matrix, at each time step,
the filter will estimate the state vector based on the IMU and landmark
measurements, as well as the covariance matrices found by IMU and
Vision-Nets. To manage computational resources effectively, the data
is divided into mini-batches of size 32. After each mini-batche is
processed, the loss value is found per \eqref{eq:loss_batch}. the
gradient of this loss with respect to the networks weights are found
and clipped to one to avoid gradient explosion. These gradients are
accumulated through mini-batches to find the gradient of the batch.
After all the mini-batches in a batch are processed, the weights are
updated using the Adam optimizer \cite{kingma2014adam}. $\mathcal{L}_{2}$
regularization \cite{loshchilov2017fixing} has been performed during
the weight training to avoid overfitting.\vspace{0.2cm}

\begin{figure}
	\centering \includegraphics[width=0.49\textwidth, height=0.25\textheight]{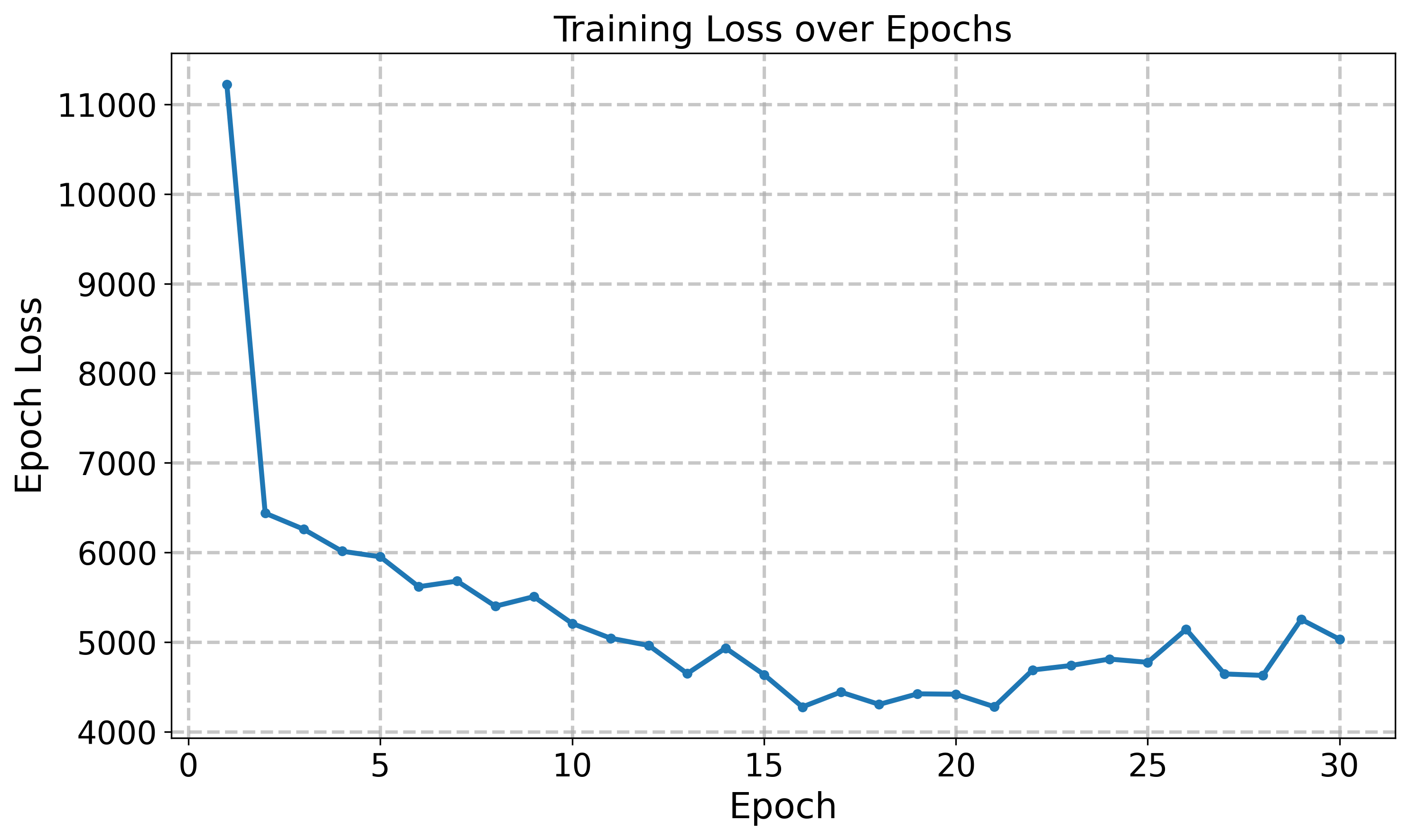}
	\caption{Training loss convergence over 30 epochs.}
	\label{fig:training_loss}
\end{figure}

\begin{figure}[h]
	\centering \includegraphics[width=0.49\textwidth, height=0.21\textheight]{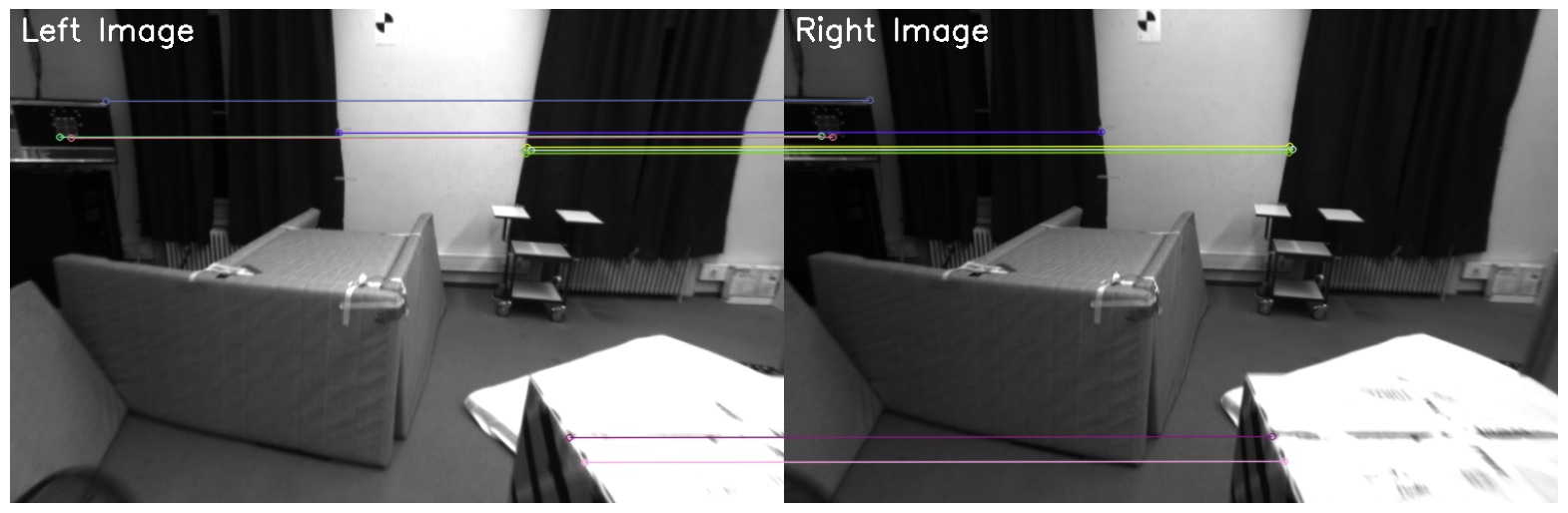}
	\caption{Matched feature points between the left and right images of a set
		of stereo image measurements using EuRoC dataset \cite{Burri25012016}.}
	\label{fig:matched_points}
\end{figure}
\begin{algorithm}[!h]
	\caption{\label{alg:Alg_training}Training Procedure}
	
	\textbf{Initialization}:
	
	\textbf{\hspace*{0.2cm}}1: Set initial values for $\hat{x}_{0}$
	and $P_{0}$.
	
	\textbf{\hspace*{0.2cm}}2: Create mini-batches of size 32 $u_{k-11:k-1}$,
	$z_{k}$, and $x_{k}$.
	
	\textbf{For} each $i$-th epoch:
	
	\textbf{\hspace*{0.2cm}}3: Initialize $\text{Gradient}\leftarrow0$.
	
	\textbf{\hspace*{0.2cm}For} each $j$-th mini-batch:
	
	\textbf{\hspace*{0.4cm}}4: Initialize $\text{Loss}_{j}^{\text{mini-batch}}\leftarrow0$.
	
	\textbf{\hspace*{0.4cm}}5: Compute parameter estimates (see \eqref{eq:IN}
	and \eqref{eq:VN}):
	
	\textbf{\hspace*{0.4cm}}$\begin{cases}
		\gamma_{1:12}^{\text{mini-batch}} & \leftarrow\text{IMUNet}(\text{mini-batch},W_{IN})\\
		\gamma_{13}^{\text{mini-batch}} & \leftarrow\text{VisionNet}(\text{mini-batch},W_{VN})
	\end{cases}$
	
	\textbf{\hspace*{0.4cm}}6: Calculate covariance scaling (see \eqref{eq:c_k adap}
	and \eqref{eq:noise elems final}):
	
	\textbf{\hspace*{0.4cm}}$Cov^{\text{mini-batch}}\leftarrow\overline{Cov}\cdot10^{\upsilon\tanh(\gamma^{\text{mini-batch}})}$
	
	\textbf{\hspace*{0.4cm}For} each data point in mini-batch (Sections
	\ref{sec:batch_pred} and \ref{sec:update}):
	
	\textbf{\hspace*{0.6cm}}7: $\hat{x}_{k|k-1},P_{k|k-1}\leftarrow\text{Predict}(\text{DataPoint},Cov_{1:12}^{\text{DataPoint}})$ 
	
	\textbf{\hspace*{0.6cm}}$\hat{x}_{k|k},P_{k|k}\leftarrow\text{Update}(\hat{x}_{k|k-1},z_{k},Cov_{13}^{\text{DataPoint}})$
	
	\textbf{\hspace*{0.6cm}}8: Store current state and covariance estimates.
	
	\textbf{\hspace*{0.4cm}End For}
	
	\textbf{\hspace*{0.4cm}}9: Compute loss for the mini-batch see \eqref{eq:loss_batch} 
	
	\textbf{\hspace*{0.4cm}}$\text{Loss}^{\text{mini-batch}}\leftarrow\text{Loss}(\hat{x}^{\text{mini-batch}},x^{\text{mini-batch}})$
	
	\textbf{\hspace*{0.4cm}}10: Compute and clip gradients:
	
	\textbf{\hspace*{0.4cm}}$\begin{cases}
		\text{Gradient}^{\text{mini-batch}} & \leftarrow\frac{\partial\text{Loss}^{\text{mini-batch}}}{\partial W_{\text{models}}}\\
		\text{Gradient}^{\text{mini-batch}} & \leftarrow\max(\text{Gradient}^{\text{mini-batch}},1)
	\end{cases}$
	
	\textbf{\hspace*{0.4cm}}11: Accumulate gradient: 
	
	\textbf{\hspace*{0.4cm}}$\text{Gradient}\leftarrow\text{Gradient}+\text{Gradient}^{\text{mini-batch}}$
	
	\textbf{\hspace*{0.2cm}End For}
	
	\textbf{\hspace*{0.2cm}}12: Update model weights using ADAM optimizer: 
	
	\textbf{\hspace*{0.2cm}}$W_{\text{models}}\leftarrow\text{ADAM}(\text{Gradient},W_{\text{models}})$
	
	\textbf{\hspace*{0.2cm}}13: Reset gradient: $\text{Gradient}\leftarrow0$
	
	\textbf{End For}
\end{algorithm}

\begin{figure*}[!t]
	\centering{}\includegraphics[width=0.8\textwidth, height=0.37\textheight]{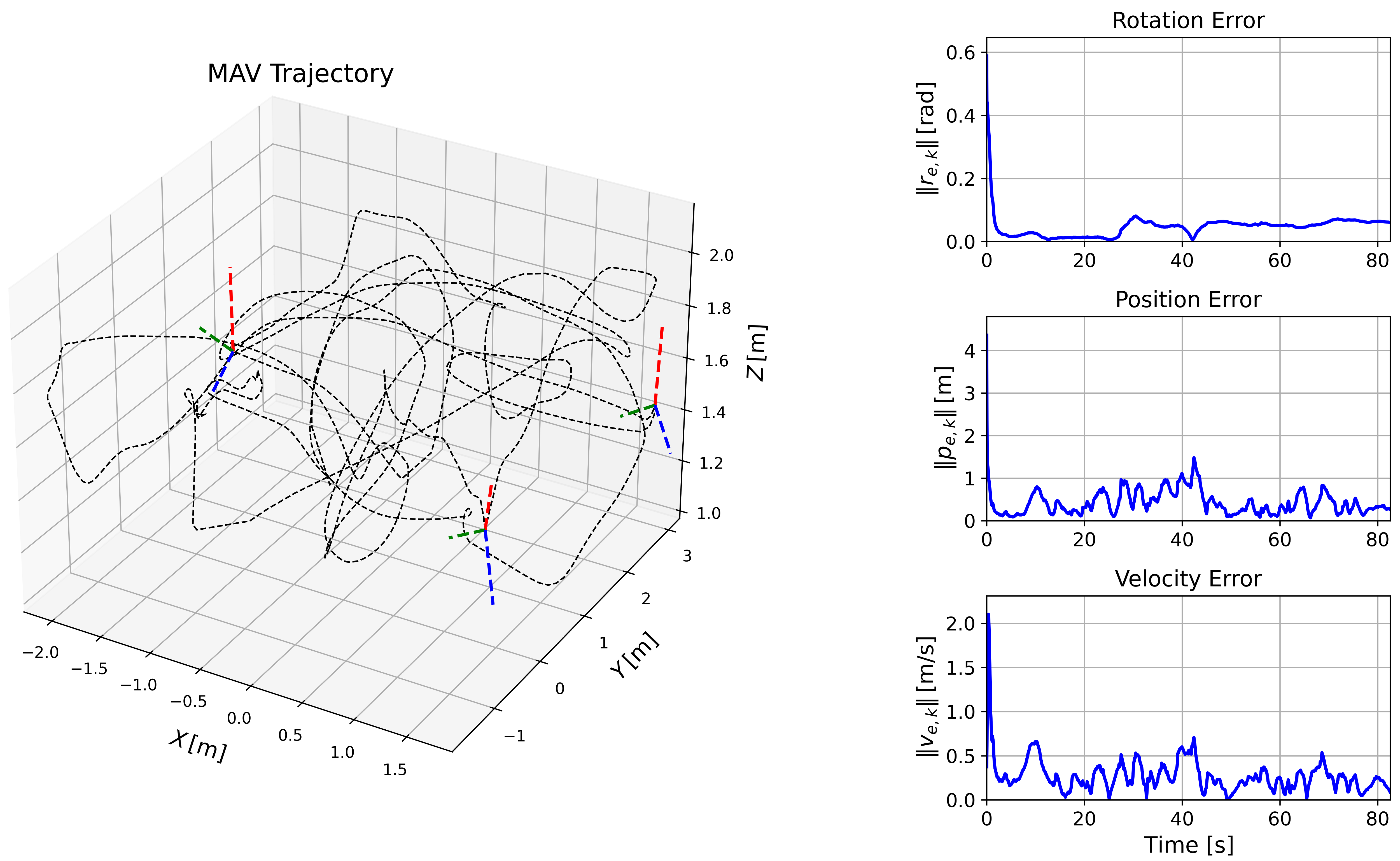}\caption{\label{fig:summary}Validation results of quaternion-based DeepUKF-VIN:
		The algorithm is evaluated using the V1\_02\_medium EuRoC dataset.
		On the left, the MAV trajectory along with three sample orientations
		in 3D space is displayed. On the right, the magnitudes of the orientation
		(top), position (middle), and velocity (bottom) vectors over time
		are illustrated.}
\end{figure*}

The IMU and Vision-Networks have 27,276 and 2,901,089 parameters,
respectively, with the weights in \eqref{eq:loss_batch} set to $w_{q}=1000$,
$w_{p}=600$, and $w_{v}=100$. The quaternion-based UKF involves
eigenvalue decomposition in \eqref{eq:weighted_average} and the use
of singular value decomposition (SVD) for computing the matrix square
root in \eqref{eq:Sigma_QNUKF}. These operations introduce significant
challenges in computing the loss gradient, particularly in step 10
of Algorithm \ref{alg:Alg_training}. To address these challenges,
we employed an EKF as the filter during the model training phase.
This substitution simplifies gradient computation considerably, thereby
enhancing the training efficiency. Despite this modification, we hypothesize
that the model can learn the optimal covariance matrices corresponding
to sensor uncertainties independently of the filter type used. This
hypothesis will be further examined in Section \ref{sec:validation}.
Given these considerations, the implementation was carried out using
PyTorch for handling the neural network components and for orientation
calculations \cite{ravi2020pytorch3d}. The models were trained over
30 epochs, during which the loss function converged to its minimum.
The convergence behaviour is illustrated in Fig. \ref{fig:training_loss}.

The measurement function is implemented by detecting 2D feature points
in each available vision dataset using the KLT algorithm. An example
of the results of this step is visualized in Fig. \ref{fig:matched_points}.
Considering the matched 2D points in the stereo images and the camera
calibration data, the feature points in the world coordinate frame
$\{\mathcal{W}\}$ are computed using triangulation \cite{hartley2003multiple}.
These computed points serve as the measurement values in this problem.
In summary, the DLAM-equipped UKF-VIN algorithm, named quaternion-based
DeepUKF-VIN, is illustrated in Fig. \ref{fig:flowchart}. The proposed
algorithm leverages IMU-Net and Vision-Net, as described in Sections
\ref{sec:IMU-Net} and \ref{sec:Vision-Net}, respectively, to compute
the covariance matrices of the UKF-VIN, as discussed in Section \ref{sec:QNUKF}.
These components are integrated to enhance the accuracy and performance
of the algorithm. The training algorithm is summarized in Algorithm
\ref{alg:Alg_training}.

\section{Experimental Validation\label{sec:validation}}
\begin{figure}[h!]
	\centering\includegraphics[width=1\linewidth]{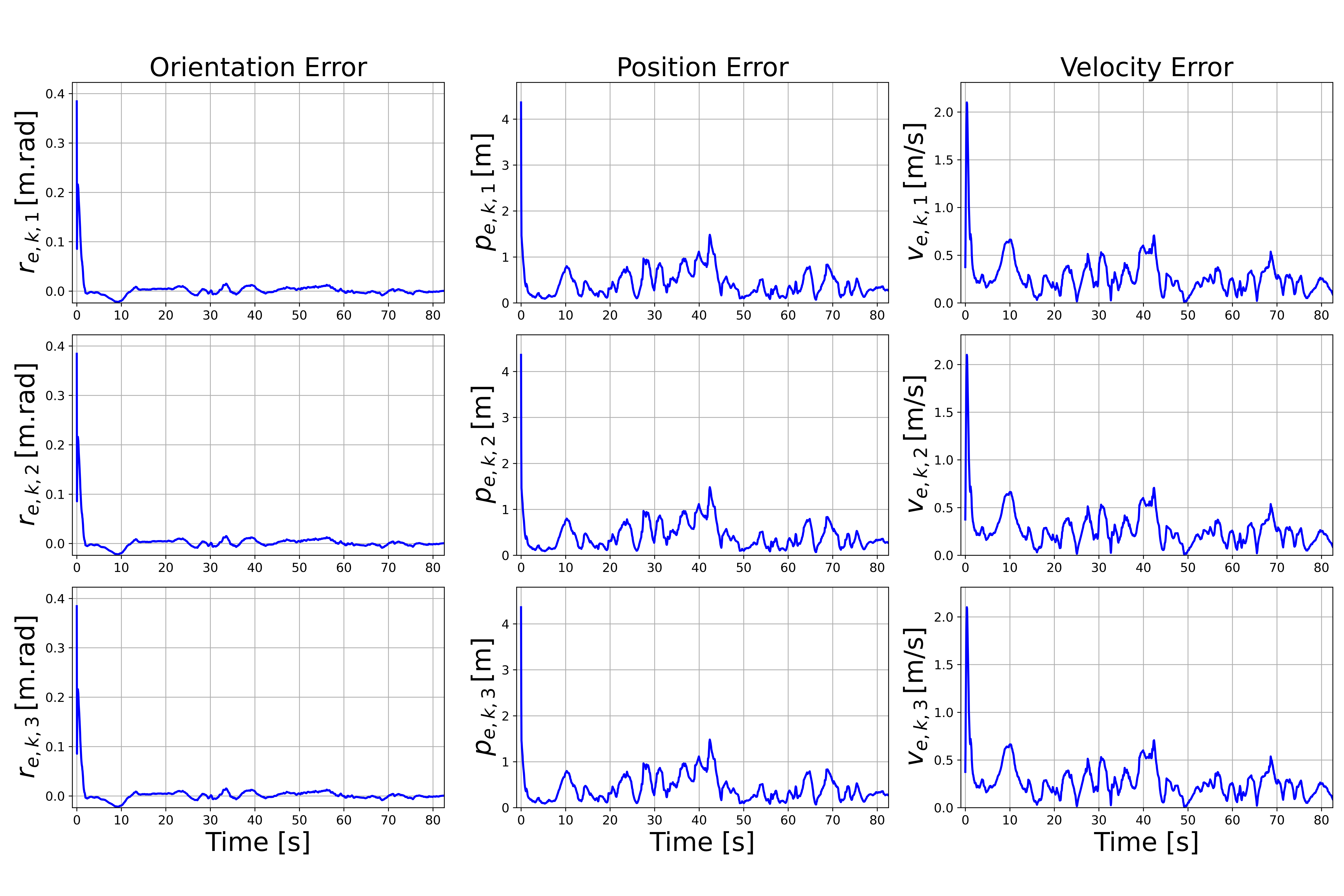}
	\caption{\label{fig:v1_ukf_components}Components of the orientation (left),
		position (middle), and velocity (right) estimation error vectors in
		the V1\_02\_medium EuRoC dataset experiment using quaternion-based
		DeepUKF-VIN.}
\end{figure}

To validate the effectiveness of quaternion-based DeepUKF-VIN, the
algorithm is tested using the real-world V1\_02\_medium EuRoC dataset
\cite{Burri25012016}. For video of the experiment, visit the following
\href{https://youtu.be/japfpySxilA}{link}. The dataset trajectory,
as well as the magnitudes of orientation, position, and velocity errors
defined in \eqref{eq:estimation_errors} over time, are visualized
in Fig. \ref{fig:summary}. The errors converge rapidly to near-zero
values despite initially high magnitudes, demonstrating quaternion-based
DeepUKF-VIN's efficacy. To further examine the results, the individual
components of each estimation error are visualized in Fig. \ref{fig:v1_ukf_components}.
It can be observed that all error components converge to near-zero
values promptly, further underscoring the effectiveness of the proposed
filter.

To investigate the effectiveness of the proposed IMU and Vision-Nets,
the quaternion-based DeepUKF-VIN was compared to its non-deep counterpart,
UKF-VIN, and another Kalman-type filter with a learning component,
the DeepEKF. The DLAM algroithm were evaluated in two environments,
V1\_02\_medium and V2\_02\_medium, which are subsets of the EuRoC
dataset \cite{Burri25012016}. Note that V2\_02\_medium dataset was
not utilized during training or validation phases. The dataset was
recorded using an Aptina MT9V034 global shutter camera, which captured
stereo images at a rate of 20 Hz. Additionally, an ADIS16448 sensor
was employed to capture IMU data at 200 Hz, while ground truth data
was recorded at 200 Hz using the Vicon motion capture system. To ensure
a fair comparison, all filters were configured with the same nominal
covariances as DeepUKF-VIN. Furthermore, in each environment, all
filters were initialized with the same state vector and covariance
matrix. The loss values, as defined in \eqref{eq:loss_batch}, for
the aforementioned filters in both experiments are presented in Table
\ref{tab:loss_comp}. The proposed DeepUKF-VIN outperformed both its
non-deep counterpart (UKF-VIN) and the DeepEKF in terms of loss values
across both experiments. The DeepEKF was exposed to data from the first experiment, while the DeepUKF-VIN was never trained on either experiment. Thus, both experiments were entirely novel to the DeepUKF-VIN. To further examine these experiments, the MSE values of the orientation,
position, and velocity estimation errors are tabulated in Table \ref{tab:all_comp}.
It can be observed from Table \ref{tab:all_comp} that across both
experiments and all components, quaternion-based DeepUKF-VIN consistently
outperformed the non-deep UKF-VIN and DeepEKF. Specifically, the DeepUKF-VIN
yielded lower MSE values across all tested experiments and components,
demonstrating its superior performance in orientation, position, and
velocity estimation.

\begin{table}[h]
	\centering{}\caption{\label{tab:loss_comp}Loss Value Comparison of DeepUKF-VIN against
		UKF-VIN and DeepEKF.}
	\begin{tabular}{lll}
		\hline 
		\textbf{Filter} & \textbf{V1\_02\_medium} & \textbf{V2\_02\_medium}\tabularnewline
		\hline 
		\hline 
		\multirow{1}{*}{DeepEKF} & 1918 & 834\tabularnewline
		\hline 
		\multirow{1}{*}{UKF-VIN} & 132 & 251\tabularnewline
		\hline 
		\multirow{1}{*}{DeepUKF-VIN} & 88 & 250\tabularnewline
		\hline 
	\end{tabular}
\end{table}

\begin{table}[h]
	\centering{}\caption{\label{tab:all_comp}Components MSEs for the two filters in each experiment}
	\begin{tabular}{llcc}
		\hline 
		\textbf{Filter} & \textbf{MSE Element} & \textbf{V1\_02\_medium} & \textbf{V2\_02\_medium}\tabularnewline
		\hline 
		\hline 
		\multirow{3}{*}{DeepEKF} & Orientation & 1.572 & 0.0544\tabularnewline
		& Position & 9.3188 & 1.1228\tabularnewline
		& Velocity & 7.5663 & 0.9558\tabularnewline
		\hline 
		\multirow{3}{*}{UKF-VIN} & Orientation & 0.0015 & 0.0026\tabularnewline
		& Position & 0.0929 & 0.3070\tabularnewline
		& Velocity & 0.0509 & 0.1319\tabularnewline
		\hline 
		\multirow{3}{*}{DeepUKF-VIN} & Orientation & 0.0008 & 0.0080\tabularnewline
		& Position & 0.0806 & 0.3011\tabularnewline
		& Velocity & 0.0282 & 0.0914\tabularnewline
		\hline 
	\end{tabular}
\end{table}

\section{Conclusion \label{sec:Conclusion}}

In this paper, we proposed an adaptively-tuned Deep Learning Unscented
Kalman Filter for 3D Visual-Inertial Navigation (DeepUKF-VIN) to estimate
the orientation, position, and velocity of a vehicle with six degrees
of freedom (6-DoF) in three-dimensional space. By effectively addressing
kinematic nonlinearities through a quaternion-based framework, the
algorithm mitigates numerical instabilities commonly associated with
Euler-angle representations. DeepUKF-VIN integrates data from a 6-axis
Inertial Measurement Unit (IMU) and stereo cameras, achieving robust
navigation even in GPS-denied environments. The Deep Learning-based
Adaptation Mechanism (DLAM) dynamically adjusts noise covariance matrices
based on sensor data, improving estimation accuracy by responding
adaptively to varying conditions. Evaluated with real-world data from
low-cost sensors operating at low sampling rates, DeepUKF-VIN demonstrated
stability and rapid error attenuation. Comparative testing across
two experimental setups consistently showed that DeepUKF-VIN outperformed
the standard Unscented Kalman Filter (UKF) in all key navigation components.
These results underscore the algorithm's superior adaptability, efficacy,
and robustness in practical scenarios, validating its potential for
accurate and reliable 3D navigation.

Future work could explore the application of the proposed DLAM to
other Kalman-type and non-Kalman-type filters developed for the VIN
problem. Given that the proposed DLAM was trained using the Extended
Kalman Filter (EKF) and validated with the UKF, it is reasonable to
hypothesize that integrating DLAM with other algorithms may yield
similar benefits. Furthermore, the vision-based component of the proposed
DLAM could be adapted for alternative sensor inputs, such as Light
Detection and Ranging (LiDAR) and Sound Navigation and Ranging (SONAR),
with minimal modifications. Such adaptations have the potential to
enhance the performance of algorithms relying on these sensor technologies.

\balance

\bibliographystyle{IEEEtran}
\bibliography{bib_QUKF_ACC}

\end{document}